\documentclass[10pt]{article} % For LaTeX2e
\usepackage[accepted]{tmlr}
\usepackage{amsmath,amsfonts,bm}

\def\eqref#1{equation~\ref{#1}}
\def\1{\bm{1}}

\DeclareMathAlphabet{\mathsfit}{\encodingdefault}{\sfdefault}{m}{sl}
\SetMathAlphabet{\mathsfit}{bold}{\encodingdefault}{\sfdefault}{bx}{n}

\DeclareMathOperator*{\argmin}{arg\,min}

\usepackage{hyperref}
\hypersetup{
    plainpages=false,
    colorlinks,
    linkcolor={red!50!black},
    citecolor={blue!50!black},
    urlcolor={blue!80!black}
}
\usepackage{url}

\usepackage{amsmath}
\usepackage{amssymb}
\usepackage{mathtools}
\usepackage{amsthm}

\usepackage[capitalize,noabbrev]{cleveref}

\theoremstyle{plain}
\newtheorem{theorem}{Theorem}[section]
\newtheorem{proposition}[theorem]{Proposition}
\newtheorem{lemma}[theorem]{Lemma}
\newtheorem{corollary}[theorem]{Corollary}
\theoremstyle{definition}

\theoremstyle{remark}
\newtheorem{remark}[theorem]{Remark}

\usepackage{graphicx}
\usepackage{subcaption}
\usepackage{mwe}
\usepackage{wrapfig}
\usepackage{booktabs}
\usepackage{multirow}
\usepackage{makecell}
\usepackage{wrapfig}
\usepackage{xspace}
\usepackage{enumitem}
\usepackage{siunitx}

\newcommand{\bdre}{\textsc{BDRE}\xspace}
\newcommand{\COB}{\textsc{Multinomial Logistic Regression based Density Ratio Estimator}\xspace}
\newcommand{\cob}{\textsc{MDRE}\xspace}

\newcommand{\cauchydist}{\mathcal{C}\text{auchy}}
\newcommand{\optmodel}{{\theta^\ast}}

\makeatletter
\newcommand{\printfnsymbol}[1]{%
  \textsuperscript{\@fnsymbol{#1}}%
}

\title{Estimating the Density Ratio between Distributions with High Discrepancy using Multinomial Logistic Regression} 

\author{\name Akash Srivastava\thanks{equal contribution} \email akashsri@mit.edu \\
      \addr MIT-IBM Watson AI Lab and IBM Research
      \AND
      \name Seungwook Han$^*$ \email swhan@mit.edu \\
      \addr MIT
      \AND
      \name Kai Xu \email xuk@amazon.com\\
      \addr Amazon\AND
      \name Benjamin Rhodes \email ben.rhodes@ed.ac.uk\\
      School of Informatics, University of Edinburgh\AND
      \name Michael U. Gutmann$^*$ \email michael.gutmann@ed.ac.uk\\
      School of Informatics, University of Edinburgh
      }

\def\month{03}  % Insert correct month for camera-ready version
\def\year{2023} % Insert correct year for camera-ready version
\def\openreview{\url{https://openreview.net/forum?id=jM8nzUzBWr}} % Insert correct link to OpenReview for camera-ready version 

\begin{document}

\maketitle

\begin{abstract}

Functions of the ratio of the densities $p/q$ are widely used in machine learning to quantify the discrepancy between the two distributions $p$ and $q$. For high-dimensional distributions, binary classification-based density ratio estimators have shown great promise. However, when densities are \emph{well separated}, estimating the density ratio with a binary classifier is challenging. In this work, we show that the state-of-the-art density ratio estimators perform poorly on \emph{well separated} cases and demonstrate that this is due to distribution shifts between training and evaluation time. We present an alternative method that leverages multi-class classification for density ratio estimation and does not suffer from distribution shift issues. The method uses a set of auxiliary densities $\{m_k\}_{k=1}^K$ and trains a multi-class logistic regression to classify the samples from $p, q$ and $\{m_k\}_{k=1}^K$ into $K+2$ classes. { We show that if these auxiliary densities are constructed such that they overlap with $p$ and $q$, then a multi-class logistic regression 
allows for estimating $\log p/q$ on the domain of any of the $K+2$ distributions and resolves the distribution shift problems of the current state-of-the-art methods.}
We compare our method to state-of-the-art density ratio estimators on both synthetic and real datasets and demonstrate its superior performance on the tasks of density ratio estimation, mutual information estimation, and representation learning. Code: \url{https://www.blackswhan.com/mdre/}
\end{abstract}

\section{Introduction}
\label{sec:intro}
Quantification of the discrepancy between two distributions underpins a large number of machine learning techniques.
For instance, distribution discrepancy measures known as $f$-divergences \citep{csiszar1964}, which are defined as expectations of convex functions of the ratio of two densities, are ubiquitous in many domains of supervised and unsupervised machine learning. Hence, density ratio estimation is often a central task in generative modeling, mutual information and divergence estimation, as well as representation learning \citep{sugiyama2012density,nce,gan,fgan,veegan,belghazi2018mine,infonce,gram}. However, in most problems of interest, estimating the density ratio by modeling each of the densities separately is significantly more challenging than directly estimating their ratio for high dimensional densities
\citep{sugiyama2012density}. Hence, direct density ratio estimators are often employed in practice.
% , usually given samples from the pair of densities only.

\begin{figure}[]
    \centering
    % \begin{subfigure}[b]{.15\textwidth}
    %     \centering
    %     \includegraphics[height=3.5cm,width=2.5cm]{fig/1d/boundary/1D_demo_cauchy_hist.png}
    %     \caption[]%
    %     {{\small Distributions p, q and m.}}
    %     \label{fig:hist_pqm} height=2.5cm,width=2.5cm
    % \end{subfigure}
    \begin{subfigure}[b]{.245\textwidth}
        \centering
        \raisebox{1.0cm}{\rotatebox{90}{\tiny{probability}}}
        \includegraphics[height=3.25cm,width=3.75cm,trim={0.4cm 0 0 0},clip]{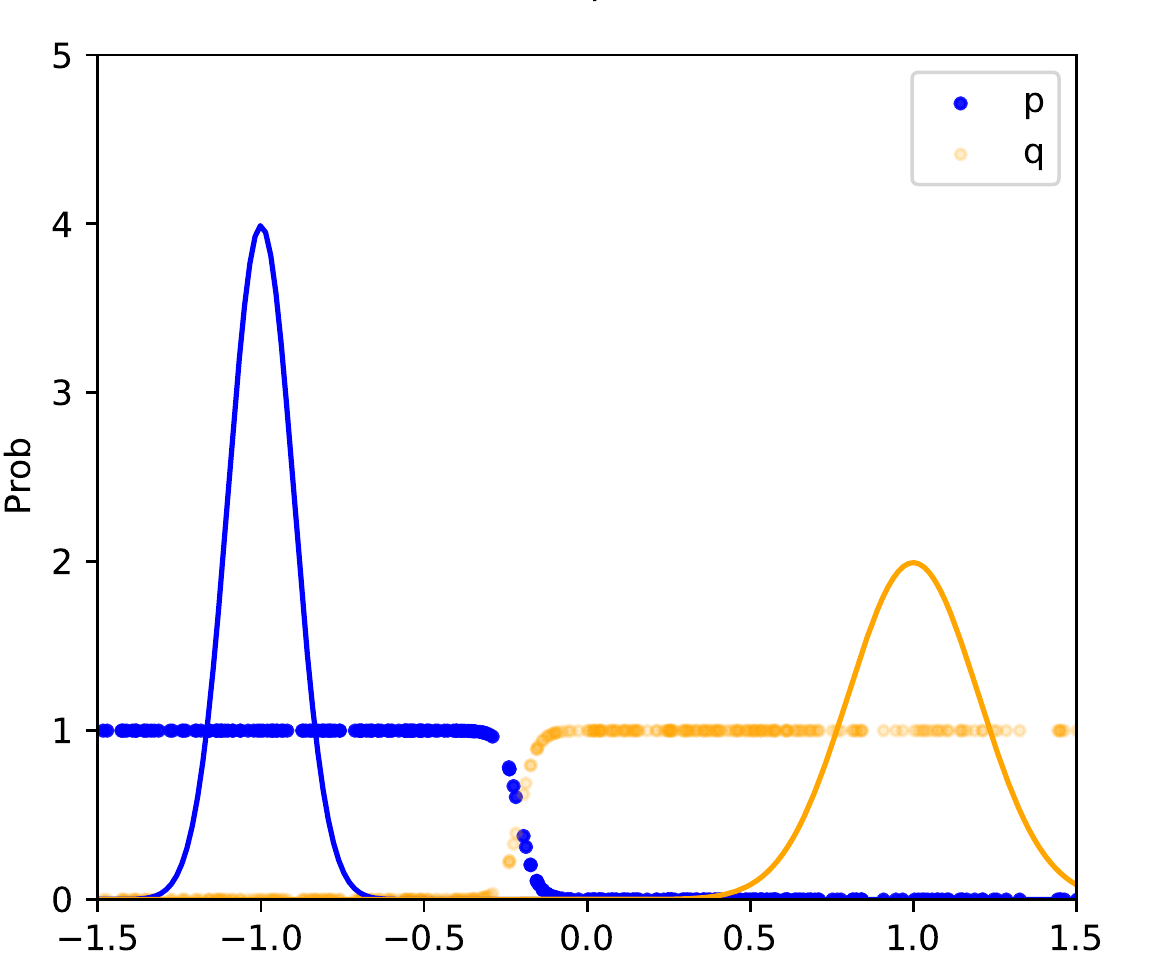}
        \caption[]
        {\small BDRE class probability}
        \label{fig:bcdre_db}
    \end{subfigure}
    \begin{subfigure}[b]{.245\textwidth}
        \centering
        \raisebox{1.25cm}{\rotatebox{90}{\tiny{log-ratio}}}
        \includegraphics[height=3.25cm,width=3.75cm]{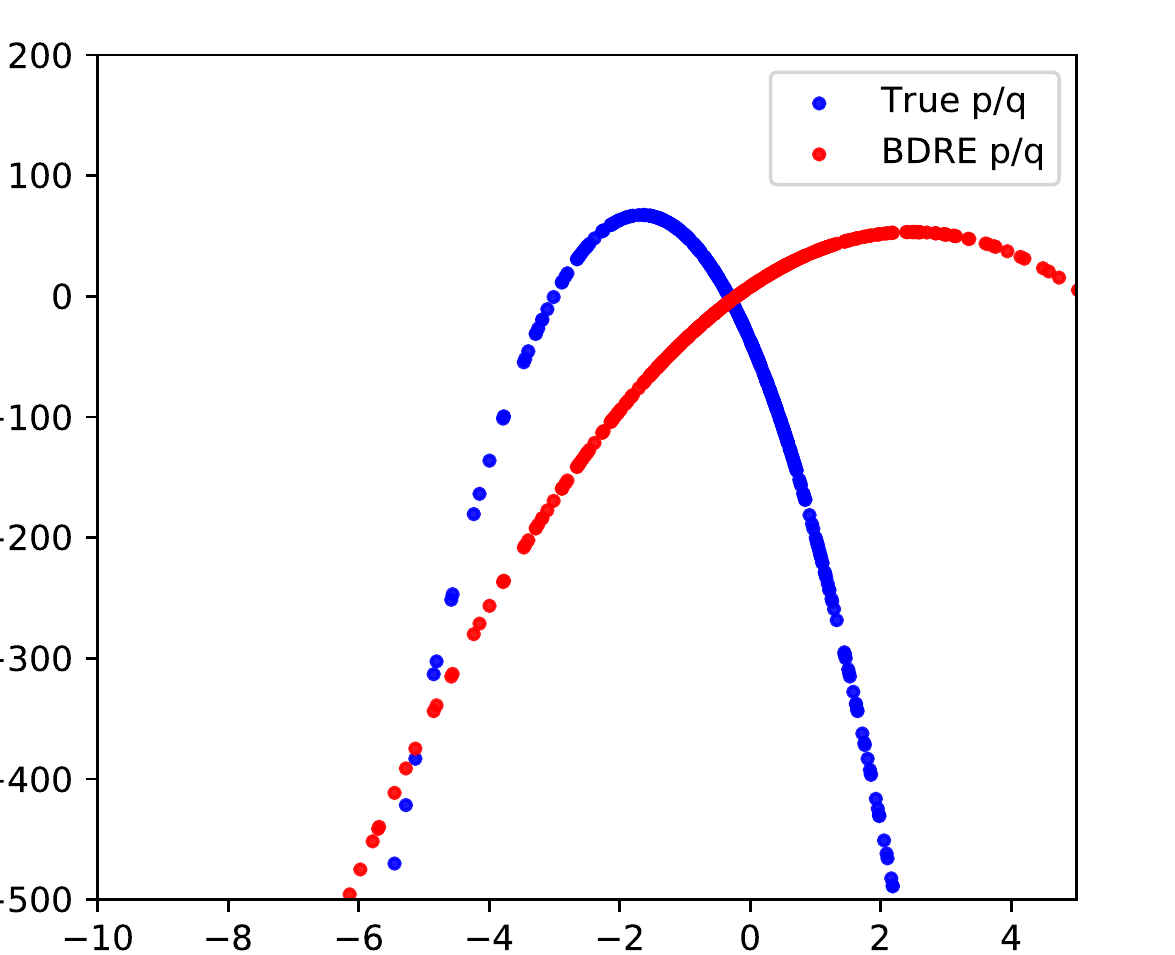}
        \caption[]
        {\small BDRE log density-ratio}
        \label{fig:bcdre_lr}
    \end{subfigure}
    \begin{subfigure}[b]{.245\textwidth}
        \centering
        \raisebox{1.0cm}{\rotatebox{90}{\tiny{probability}}}
        \includegraphics[height=3.25cm,width=3.75cm,trim={0.4cm 0 0 0},clip]{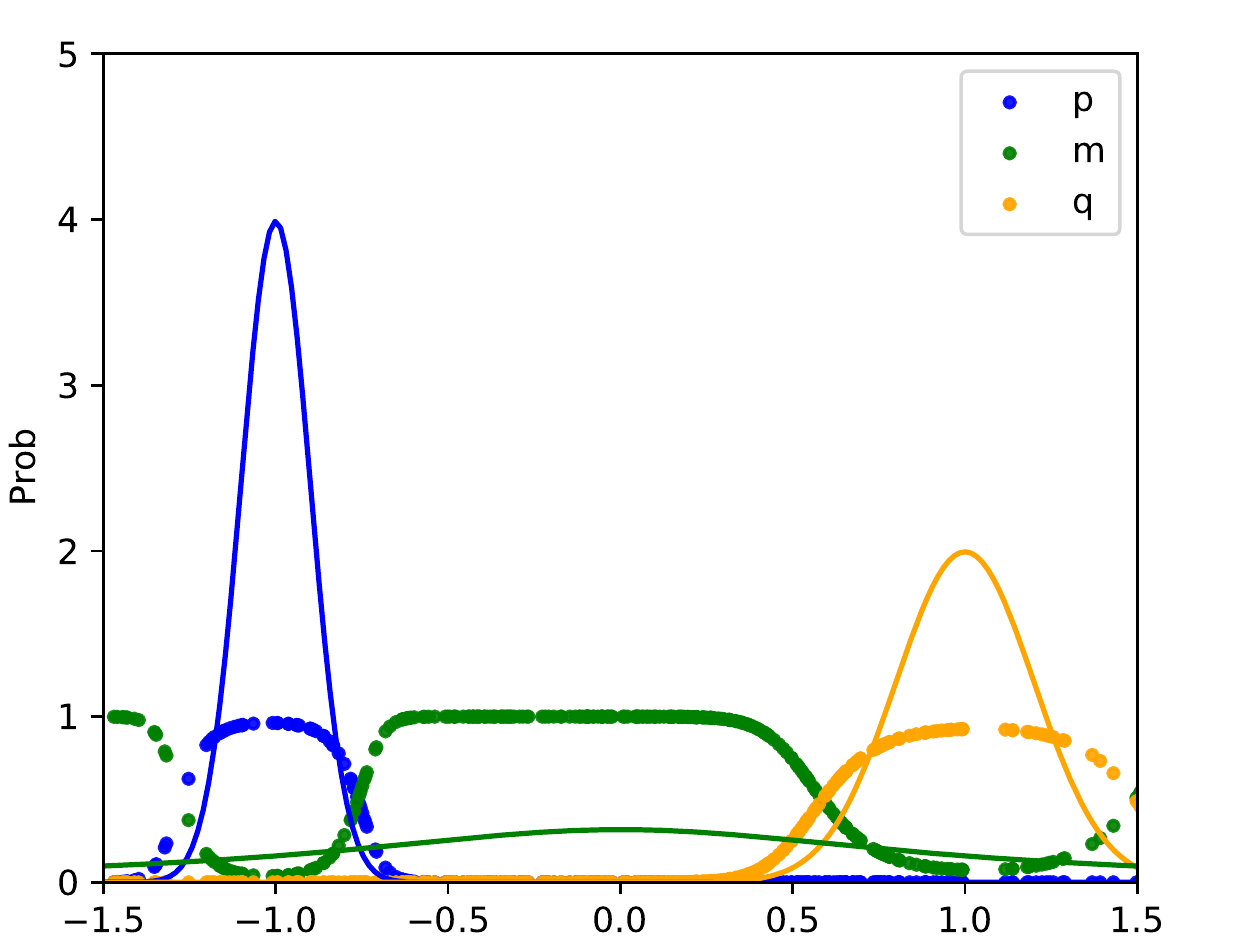}
        \caption[]
        {{\small \cob class probability}}
        \label{fig:cob_db}
    \end{subfigure}
    \begin{subfigure}[b]{.245\textwidth}
        \centering
        \raisebox{1.25cm}{\rotatebox{90}{\tiny{log-ratio}}}
        \includegraphics[height=3.25cm,width=3.75cm]{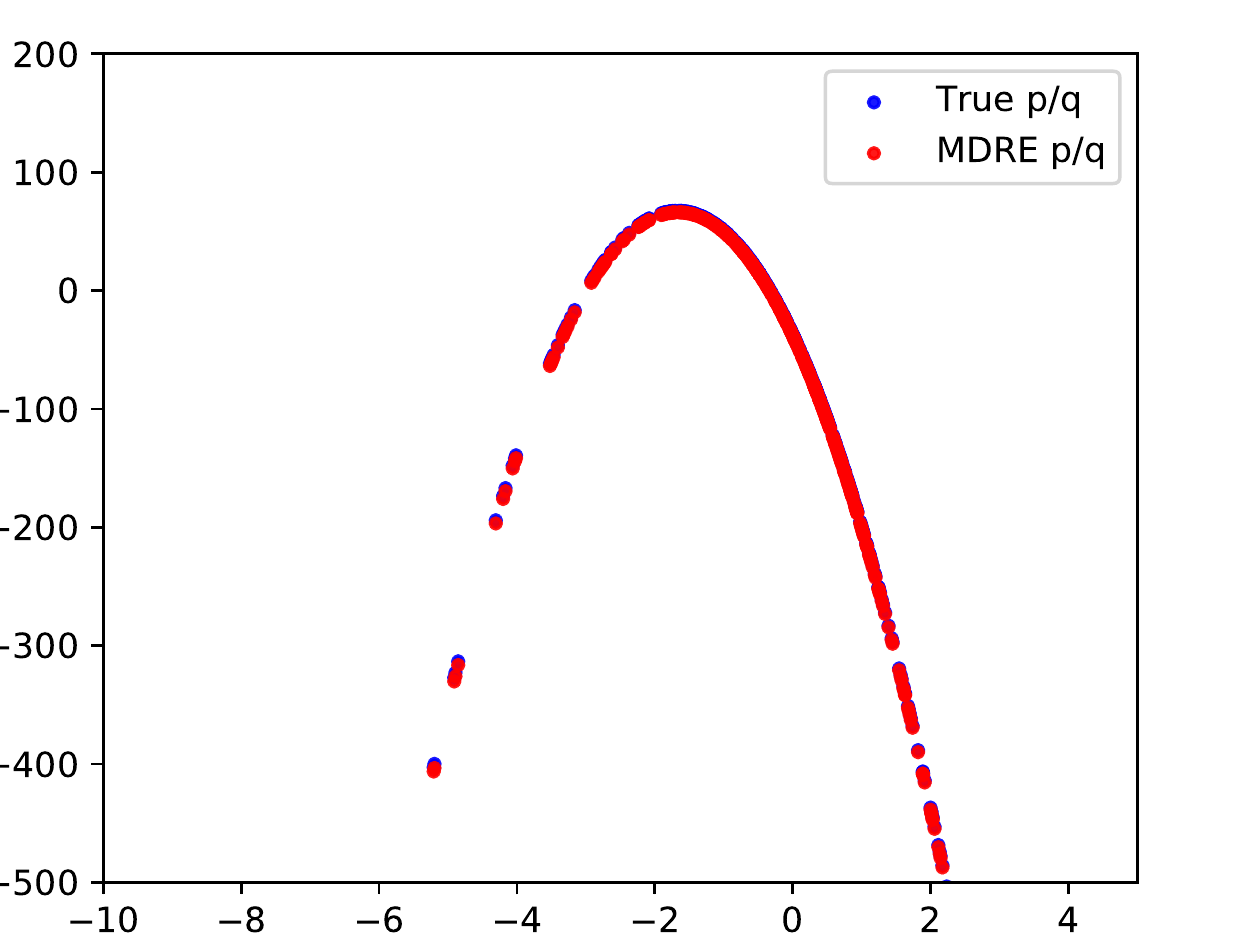}
        \caption[]
        {{\small \cob log density-ratio}}
        \label{fig:cob_lr}
    \end{subfigure}
    % % \vskip\baselineskip
    \caption{BDRE vs proposed \cob on estimation of $\log$ density ratio where $p=\mathcal{N}(-1,0.1)$ and $q=\mathcal{N}(1,0.2)$. For \cob, the auxillary distribution $m$ is Cauchy $\mathcal{C}(0,1)$. Plots (a) and (c) show the class probabilities $P(Y|x)$ learned for BDRE and \cob respectively overlayed on the plots of $p$, $q$ and $m$. Plots (b) and (d) show the estimated log density-ratio by BDRE and \cob respectively. Using auxiliary distribution $m$ allows \cob to better estimate the log density-ratio.}
    \label{fig:boundary_and_log_ratio}
\end{figure}

One of the most commonly used density ratio estimators (DRE) utilizes binary classification via logistic regression (BDRE). Once trained to discriminate between the samples from the two densities, BDREs have been shown to estimate the ground truth density ratio between the two densities \citep[e.g.][]{nce, Gutmann2011b, sugiyama2012density,menon2016linking}.
BDREs have been tremendously successful in problems involving the minimization of the density-ratio based estimators of discrepancy between the data and the model distributions even in high-dimensional settings \citep{fgan,dcgan}. However, they do not fare as well when applied to the task of estimating the discrepancy between two distributions \emph{that are far apart or easily separable from each other}. This issue has been characterized recently as the \emph{density-chasm problem} by \citet{tre}. 
% When two distributions are easily separable, leading to an easier classification problem for BCDREs, the resulting density-ratio estimates become inaccurate. 
We demonstrate this in Figure \ref{fig:boundary_and_log_ratio} where we employ a BDRE to estimate the density ratio between two 1-D distributions, $p=\mathcal{N}(-1,0.1)$ and $q=\mathcal{N}(1,0.2)$ shown in panel (a). Since $p$ and $q$ are considerably far apart from each other, solving the classification problem is relatively simple as illustrated by the visualization of the decision boundary of the BDRE. However, as shown in panel (b), even in this simple setup, BDRE completely fails to estimate the ratio. 
\citet{kato2021non} have also confirmed that most DREs, especially those implemented with deep neural networks, tend to overfit to the training data in some way when faced with the density-chasm problem. { Since BDRE-based plug-in estimators are often used in many high-dimensional tasks such as mutual information estimation, representation learning, energy-based modeling, co-variate-shift resolution, and importance sampling \citep{tre, choi2021density, choi2021featurized, sugiyama2012density}, resolving density-chasm is an important problem of high practical relevance.}

A recently introduced solution to the density-chasm problem, telescopic density-ratio estimation \citep[TRE; ][]{tre}, tackles it by replacing the easier-to-classify, original logistic regression problem, by a \textit{set} of harder-to-classify logistic regression problems.
In short, TRE constructs a set of $K$ auxiliary distributions ($\{m_k\}_{k=1}^K$) to bridge the two target distributions ($p=:m_0$ and $q=:m_{K+1}$) of interest and then trains a set of $K+1$ BDREs on every pair of \emph{consecutive distributions} ($m_{k-1}$ and $m_k$ for $k = 1, \dots, K$), which are assumed to be close enough (i.e. not easily separable) for BDREs to work well.
After that, an overall density ratio estimate is obtained by taking the cumulative (telescopic) product of all individual estimates.

In this work, we argue that the aforementioned solution to the density chasm problem has an inherent issue of \emph{distribution shift} that can lead to significant inaccuracies in the final density ratio estimation.
Notice that the $i$-th BDRE in the chain of BDREs that TRE constructs is only trained on the samples from distributions $m_{i}$ and $m_{i+1}$. However, post-training, it is typically evaluated on regions where the distributions from the original density ratio estimation problem (i.e.\ $p$ and $q$) have non-negligible mass. If the high-probability regions of $p$, $q$ and the auxiliary distributions $m_i$ do not overlap, the training and evaluation distributions for the $i$-th BDRE are different. Because of this distribution shift between training and evaluation, the overall density ratio estimation can end up being inaccurate (see Figure~\ref{fig:dist_shift-tre} and Section~\ref{sec:dist_shift-tre} for further details).
We here provide another solution to the density-chasm problem that avoids this distribution shift. 

We present \COB (\cob), a novel method for density ratio estimation that solves the density-chasm problem without suffering from distribution shift. This is done by using auxiliary distributions and \emph{multi-class classification}. \cob replaces the easy binary classification problem with a \emph{single} harder multi-class classification problem. 
\cob first constructs a set of $K$ auxiliary distributions $\{m_k\}_{k=1}^K$ that overlap with $p$ and $q$ and then uses multi-class logistic regression on the $K+2$ distributions to obtain a density ratio estimator of $\log p/q$. We will show that the multi-class classification formulation avoids the distribution shift issue of TRE.

The key contributions of this work are as follows:
\begin{enumerate}
    \item We study the state-of-the-art solution to the density-chasm problem \citep[TRE;][]{tre} and identify its limitations arising from distribution shift. We illustrate that this inherent issue can significantly degrade its density ratio estimation performance. 
    \item We formally establish the link between multinomial logistic regression and density ratio estimation and propose a novel method (\cob) that uses auxiliary distributions to train a multi-class classifier for density ratio estimation. { \cob resolves the aforementioned distribution shift issue by construction and effectively tackles the density chasm problem.}
    \item We construct a comprehensive evaluation protocol that significantly extends on benchmarks used in prior works. We conduct a systematic empirical evaluation of the proposed approach and demonstrate the superior performance of our method on a number of synthetic and real datasets. Our results show that \cob is often markedly better than the current state-of-the-art of density ratio estimation on tasks such as $f$-divergence estimation, mutual information estimation, and representation learning in high-dimensional settings. 
\end{enumerate}

% \input{s2-back-old}
% NOTE There is no background section any more
% \input{s3-related-old}
\section{Related Work}
\label{sec:related_work}

% \paragraph{TRE \& DRE-$\infty$}
Telescopic density-ratio estimation \citep[TRE, ][]{tre} uses a two step, divide-and-conquer strategy to tackle the density-chasm problem. In the first step, they construct $K$ \emph{waymark} distributions $\{m_k\}_{k=1}^K$ by gradually transporting samples from $p$ towards samples from $q$. Then, they train $K$ BDREs, one for each consecutive pair of distributions. This allows for estimating the ratio $r_{p/q}$ as the product of $K+1$ BDREs, $r_{p/q} := \frac{p}{q} = \frac{p}{m_1} \times \dots \times \frac{m_K}{q}$. 
\cite{tre} introduced two schemes for creating waymark distributions that ensure that consecutive pairs of distributions are packed \emph{closely enough} so that none of the $K+1$ BDREs suffer from the density-chasm issue. Hence, TRE addresses the density-chasm issue by replacing the ratio between $p$ and $q$ with a product of $K+1$ intermediate density ratios that, by design of the waymark distribution, should not suffer from the density-chasm problem. In a new work, \citet{choi2021density} introduced $\text{DRE-}\infty$, a method that takes the number of waymark distributions in TRE to infinity and derives a limiting objective that leads to a more scalable version of TRE.

% \paragraph{F-DRE}
F-DRE is other interesting related work that comes from \citet{choi2021featurized}. F-DRE uses a FLOW-based model \citep{rezende2015variational} which is trained to project samples from a mixture of the two distributions onto a standard Gaussian. They then train a BDRE. It is easy to show that any bijective map will preserve the original density ratio $r_{p/q}$ in the feature space as the Jacobian correction term simply cancels out. However, due to the bijectivity of the FLOW map, such a method cannot bring the projected distributions any closer than the discrepancy between the original distributions. At best, the method can shift the discrepancy between the original distributions along different moments after projection. Due to this issue, we found that F-DRE did not work well for the problems we considered (see experimental results in Section \ref{sec:exp}). 
Recently, \citet{liu2021analyzing} introduced an optimization-based solution to the density-chasm problem in exponential family distributions by using (a) normalized gradient descent and (b) replacing the logistic loss with an exponential loss. 
Finally, while BDRE remains the dominant method of density ratio estimation in recent literature, prior works, such as ~\cite{bickel2008multi} and \cite{nock2016scaled}, have studied multi-class classifier-based density ratio estimation for estimating ratios between a set of densities against a common reference distribution and its applications in multi-task learning. 

\subsection{TRE's performance can degrade due to training-evaluation distribution shifts}
\label{sec:dist_shift-tre}
In supervised learning, distribution shift \citep{quinonero2009dataset} occurs when the training data $(x,y) \sim p_{\text{train}}$ and the test data $(x,y)\sim p_{\text{test}}$ come from two different
distributions, i.e.\ $p_{\text{train}} \neq p_{\text{test}}$.
Common training methods, such as those used in BDRE, only guarantee that the model performs well on unseen data that comes from the same distribution as $p_{\text{train}}$.
Thus, in the case of distribution shift at test time, the model's performance degrades proportionately to the shift.
We now show that a similar distribution shift can occur in TRE when distributions $p$ and $q$ are sufficiently different. Recall that in TRE, we use BDREs to estimate $K+1$ density ratios $p/m_1, m_2/m_1, \dots, m_K/q$ that are combined in a telescopic product to form the overall ratio $p/q$. Let us denote the estimates of the $K+1$ ratios by $\hat \eta_1, \dots, \hat \eta_{K+1}$. 

Given the theoretical properties of BDRE, for any $i \in \{1,\dots,K+1\}$, $\hat \eta_i$  estimates $r_{m_{i-1}/m_i}$ over the support of $m_i$ \citep{sugiyama2012density, nce, menon2016linking}. However, in TRE, when we evaluate the target ratio $p/q$ on the supports of $p$ and $q$, we evaluate the individual $\hat \eta_i$ on domains for which we lack guarantees that they perform well. Since the overall estimator for $p/q \approx \hat \eta_1 * \dots * \hat \eta_{K+1}$ combines multiple ratio estimators, it suffers from the distribution shift issue if \emph{any} of the individual estimators' performance deteriorates. 
Thus, if the supports of $\{m_i\}_{i=1}^K$, $p$, and $q$ are different, or when the samples from $\{m_i\}_{i=1}^K$, $p$, and $q$ do not overlap well enough, the training and evaluation domains of the $\hat \eta_i$ are different and we expect the ratio estimate $\hat \eta_i$ and, in turn, the overall estimator for $p/q$ to be poor. We now illustrate this with a toy example.

\begin{figure}[t]
    \centering
    \includegraphics[width=.95\textwidth]{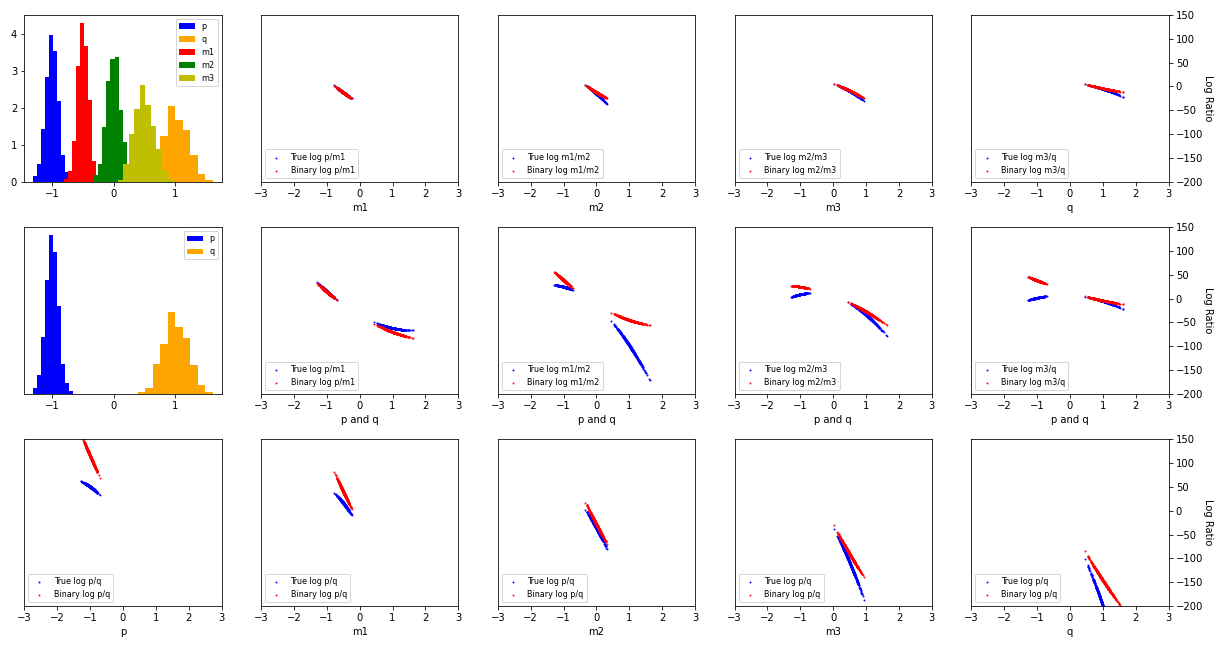}
    \caption{TRE for $p=\mathcal{N}(-1,0.1)$ and $q=\mathcal{N}(1,0.2)$ from Figure \ref{fig:boundary_and_log_ratio}. In all scatter plots, x-axis denotes the sampling distribution and y-axis denotes the log-density-ratio. The density plot in the first row shows $p$, $q$ and the 3 waymarks; the density plot in the second row shows $p$ and $q$ only. The scatter plots in the first row show individual density ratio estimators evaluated on samples from their corresponding training data (denominator density), demonstrating accurate estimation on the training set. The scatter plots in the second show  individual density ratio estimators evaluated on samples from $p$ and $q$. The estimation accuracy has degraded notably due to the train-eval distribution shift. The last row shows the performance of the overall density ratio estimator on samples from $p,m_1,m_2,m_3,q$. We see that the overall ratio estimate is significantly affected by the deterioration of the individual ratio estimates, illustrating the sensitivity of TRE to distribution shift problems in case of well-separated distributions.}
    \label{fig:dist_shift-tre}
\end{figure}

We consider estimating the density ratio between $p=\mathcal{N}(-1,0.1)$ and $q=\mathcal{N}(1,0.2)$. Since, $p$ and $q$ are well separated, we introduce three auxiliary distributions $m_1, m_2, m_3$ to bridge them, providing the waymarks that TRE needs. The auxiliary distributions $m_1, m_2, m_3$ are constructed with the \textit{linear-mixing} strategy that will be described in Section \ref{sec:constructing_m}. This setup is shown in the top-left panel of Figure~\ref{fig:dist_shift-tre}. We train 4 BDREs $\hat \eta_1, \hat \eta_2, \hat \eta_3, \hat \eta_4$ to estimate ratios $p/m_1, m_1/m_2, m_2/m_3$ and $m_3/q$ respectively.
% We evaluate the performance of individual $\hat \eta_i$ on samples from all $p$, $m_i$'s and $q$.
% Figure~\ref{fig:dist_shift-tre} shows the setup as well as the results.
We begin by showing that each of the trained BDREs estimates their corresponding density ratio accurately on their corresponding training distributions. To show this, in panels 2-5 in the first row of Figure~\ref{fig:dist_shift-tre}, we evaluate $\hat \eta_1, \hat \eta_2, \hat \eta_3, \hat \eta_4$ on samples from their respective denominator densities $m_1,m_2,m_3,q$ and plot them via a scatter plot where the x-axis is labeled with the distribution that we draw the samples from and the y-axis is the log-density ratio (red). We plot the true density ratio in blue for comparison. As evident, red and blue scatter plots overlap significantly,  indicating the individual ratio estimators are accurate on their respective denominator (training) distributions.

Next, we evaluate the BDREs $\hat \eta_1, \hat \eta_2, \hat \eta_3, \hat \eta_4$, on samples from $p$ and $q$ instead of their corresponding training distributions as before. Distributions $p$ and $q$ are shown in panel 1 of the second row in Figure~\ref{fig:dist_shift-tre}. In the rest of the panels (2-5) in the second row, estimators $\hat \eta_1, \hat \eta_2, \hat \eta_3, \hat \eta_4$ are compared to the ground-truth log-density ratios (blue) $p/m_1, m_1/m_2, m_2/m_3$ and $m_3/q$ that are also evaluated on samples from $p$ and $q$. Unlike in row 1, the estimated log-density ratios do not match the ground-truth. This reflects the training-evaluation distribution-shift issues pointed out above. We show now that this deterioration in accuracy on the level of the individual BDREs results in an deterioration of the overall performance of TRE. To this end, we first recover the TRE estimator by chaining the individually trained BDREs via a telescoping product, i.e. $\hat \eta_1 * \hat \eta_2*\hat \eta_3*\hat \eta_4$ and then evaluate it on samples from all the 5 distributions $p, m_1,m_2,m_3,q$. The results are shown in panels 1-5 of the third row. The estimated log-density ratios (red) do not match the corresponding ground-truth log-density ratios (blue), which demonstrates that the distribution-shift in the training and evaluation distributions of the individual BDREs significantly degrades the overall estimation accuracy of TRE. Additional issues occur when both $p$ and $q$ do not have full support, as discussed in Appendix \ref{app:trevscob}.

%A side note is that we also found that in cases when both $p$ and $q$ do not have full support, even while estimating the $\frac{p}{q}$ accurately, the waymark construction schemes of TRE may render some of the intermediate ratios, theoretically, ill-defined during training; see Appendix \ref{app:trevscob} for a detailed discussion.

\section{Density Ratio Estimation using Multinominal Logistic Regression}
% \section{Density Ratio Estimation by auxiliary Densities}

We propose \COB (\cob) to tackle the density-chasm problem while avoiding the distribution shift issues of TRE.  As in TRE, we introduce a set of $K \geq 1$ auxiliary distributions $\{m_k\}_{k=1}^K$. But, in constrast to TRE, we then formulate the problem of estimating $\log p/q$ as a multi-class classification problem rather than a sequence of binary classification problems. 
We show that this change leads to an estimator that is accurate on the domain of all $K+2$ distributions and, therefore, does not suffer from distribution shift.

\looseness=-1

\subsection{Loss function}
\label{sec:loss-function}
We here establish a formal link between density ratio estimation and multinomial logistic regression. Consider a set of $C$ distributions $\{p_c\}_{c=1}^C$ and let $p_x(x) = \sum_{c=1}^C \pi_c p_c(x)$ be their mixture distribution, with prior class probabilities $\pi_c$.\footnote{In our simulations, we will use a uniform prior over the classes.}
 The multi-class classification problem then consists of predicting the correct class $Y \in\{1, \ldots, C\}$ from a sample from the mixture $p_x$. For this purpose, we consider the model 
\begin{align}
    P(Y=c|x; \theta)&= \frac{\pi_c\exp(h^c_\theta(x))}{\sum_{k=1}^C\pi_k\exp(h^k_\theta(x))},
    \label{eq:parametric-model}
\end{align}
where the $h^c_\theta(x)$, $c=1, \ldots, C$ are unnormalized log probabilities parametrized by $\theta$. We estimate $\theta$ by minimizing the negative multinomial log-likelihood (i.e. the softmax cross-entropy loss) $\mathcal{L}(\theta)$
\begin{align}
\label{eq:multinomial-loss}
    \mathcal{L}(\theta)&= - \sum_{c=1}^C \pi_c \mathbb{E}_{x\sim p_c}[\log P(Y=c|x; \theta)]=\sum_{c=1}^C \pi_c \mathbb{E}_{x\sim p_c}\bigg[-\log \pi_c - h^c_\theta +\log \sum_{k=1}^C \pi_k\exp(h^k_\theta(x)) \bigg],
\end{align}
where, in practice, the expectations are replaced with a sample average. We denote the optimal parameters by $\optmodel = \argmin_\theta \mathcal{L}(\theta)$. To ease the theoretical derivation, we consider the case where the $h^c_\theta(x)$ are parametrized in such a flexible way that we can consider the above loss function to be a functional of $C$ functions $h_1,\dots , h_C$,
\begin{align}
    \mathcal{L}(h_1,\dots , h_C) &=\sum_{c=1}^C \pi_c \mathbb{E}_{x\sim p_c}\bigg[-\log \pi_c - h_c +\log \sum_{k=1}^C \pi_k\exp(h_k(x)) \bigg].
    \label{eq:functional-multinomial-loss} 
\end{align}
The following propositions shows that minimizing $\mathcal{L}(h_1, \ldots, h_C)$ allows us to estimate the log ratios between any pair of the $C$ distributions $p_c$.
\begin{proposition}
Let $\hat h_1, \ldots, \hat h_C$ be the minimizers of $\mathcal{L}(h_1,\dots , h_C)$ in \eqref{eq:functional-multinomial-loss}. Then the density ratio between $p_i(x)$ and $p_j(x)$ for any $i,j \leq C$ is given by
\begin{align}
    \label{eq:estimator}
    \log\frac{p_i(x)}{p_j(x)} = \hat h_i(x)-\hat h_j(x)
\end{align}
for all $x$ where $p_x(x)=\sum_c \pi_c p_c(x) >0$.
\label{prop:cob}
\end{proposition}
% \begin{proposition}
% Let $p_x(x) = \sum_{c=1}^C \pi_c p_c(x)$ be the mixture of the $C$ distributions $\{p_c\}_{c=1}^C$ where the $c^{\text{th}}$ entry ($\pi_c$) of the $C$-dimensional vector $\pi$ denotes the mixing coefficient for distribution $p_c$. Then, given a parametric probabilistic classifier with categorical likelihood function, that models the probability of sample $x \sim p_x$ belonging to the component distribution $p_c$ as, $P(Y=c|x; \theta)= \frac{\pi_c\exp(h^c_\theta(x))}{\sum_{k=1}^C\pi_k\exp(h^k_\theta(x))}$, the density ratio between $p_i$ and $p_j$ for any $i,j \leq C$ is given by
% \begin{align}
%     \label{eq:estimator}
%     \log\frac{p_i(x)}{p_j(x)} = h_\theta^i(x)-h_\theta^j(x).
% \end{align}
% \label{prop:cob}
% \end{proposition}
% Here, $h_\theta^c$ refers to the logit corresponding to the $c^\text{th}$ class, i.e.~the unnormalized log probabilities, 
% upon maximum likelihood training on samples from $p_x$, i.e. $\theta = \theta^\ast$ where,
% \begin{equation}\label{eq:multinomial-loss}
%     \theta^\ast = \argmin_{\theta} \mathcal{L}(\theta)= \argmax_{\theta} \sum_{c=1}^C \pi_c \mathbb{E}_{x\sim p_c}[\log P(Y=c|x; \theta)].
% \end{equation}
%
\begin{proof}
We first note that the sum of expectations $\sum_{c=1}^C \pi_c \mathbb{E}_{x\sim p_c}$ in \eqref{eq:functional-multinomial-loss} is equivalent to the expectation with respect to the mixture distribution $p_x$. Writing the expectation as an integral we obtain
\begin{align}
    \mathcal{L}(h_1,\dots , h_C)&=\sum_{c=1}^C \pi_c \mathbb{E}_{x\sim p_c}[-\log \pi_c - h_c(x)] + \int p_x(x) [\log \sum_{k=1}^C \pi_k\exp(h_k(x)) ]dx.
\end{align}
The functional derivative of $\mathcal{L}(h_1,\dots , h_C)$ with respect to $h_i$, i=1\ldots, C, equals
\begin{align}
\label{eq:partial}
    \frac{\delta \mathcal{L}}{\delta h_i} = - \pi_ip_i(x) + p_x(x)\frac{\pi_i\exp(h_i(x))}{\sum_{k=1}^C\pi_k\exp(h_k(x))}
\end{align}
for all $x$ where $p_x(x)>0$. Setting the derivative to zero gives the necessary condition for an optimum
\begin{align}
\label{eq:critical}
    \frac{\pi_ip_i(x)}{p_x(x)} = \frac{\pi_i\exp(h_i(x))}{\sum_{k=1}^C\pi_k\exp(h_k(x))}, \quad \quad i=1,\dots, C, \text{ and for all } x \text{ where } p_x(x)>0.
\end{align}
The left-hand side of \eqref{eq:critical} equals the true conditional probability $P^*(Y=i|x) = \frac{\pi_ip_i(x)}{p_x(x)}$. Hence, at the critical point, $\hat h_1,\dots, \hat h_C$ are such that $P^*(Y|X)$ is correctly estimated. From \eqref{eq:critical}, it follows that for two arbitrary $i$ and $j$, we have $(\pi_ip_i)/(\pi_jp_j)=(\pi_i\exp(\hat h_i))/(\pi_j\exp(\hat h_j))$ i.e.
\begin{align}
    \label{eq:estimator}
    \log\frac{p_i(x)}{p_j(x)} = \hat h_i(x)- \hat h_j(x)
\end{align}
for all $x$ where $p_x(x)>0$, which concludes the proof.
\end{proof}
\begin{remark}[Identifiability]
While we have $C$ unknowns $h_1,\dots,h_C$ and $C$ equations in \eqref{eq:critical}, there is a redundancy in the equations because
\begin{align*}
        \sum_{i=1}^C\frac{\pi_ip_i(x)}{p_x(x)} = \sum_{i=1}^C \frac{\pi_i\exp(h_i(x))}{\sum_{k=1}^C\pi_k\exp(h_k(x))}= \frac{\sum_{i=1}^C \pi_i\exp(h_i(x))}{\sum_{k=1}^C\pi_k\exp(h_k(x))} =1
\end{align*}
This means that we cannot uniquely identify all $h_i$ by minimising \eqref{eq:functional-multinomial-loss}. However, the difference $h_i-h_j$, for $i\neq j$, can be identified and is equal to the desired log ratio between $p_i$ and $p_j$ per \eqref{eq:estimator}. 
\end{remark}
{
\begin{remark}[Effect of parametrisation and finite sample size] In practice, we only have a finite amount of training data and the parametrisation introduces constraints on the flexibility of the model. With additional assumptions, e.g.\ that the true density ratio $\log p_i(x) -\log p_j(x)$ can be modeled by the difference of $h_\theta^i(x)$ and $h_\theta^j(x)$, we show in Appendix \ref{app:consistency} that our ratio estimator is consistent. We here do not dive further into the asymptotic properties of the estimator but focus on the practical applications of the key result in \eqref{eq:estimator}.
\end{remark}
}
Importantly, \eqref{eq:estimator} allows us to estimate $r_{p/q}$ by formulating our ratio estimation problem as a multinomial nonlinear regression problem as summarized in the following corollary.
\begin{corollary}
Let the distributions of the first two classes be $p$ and $q$, respectively, i.e.\ $p_1 \equiv p, p_2 \equiv q$, and the remaining $K$ distributions be equal to the auxiliary distributions $m_i$, i.e.\ $p_3 \equiv m_1, \ldots, p_{K+2} \equiv m_K$. Then
\begin{equation}
\label{eq:rhat-def}
%\hat{r}_{p/q}(x) = h^1_{\optmodel}(x)-h^2_{\optmodel}(x).    
\log \hat{r}_{p/q}(x) = \hat h_1(x)- \hat h_2(x).    
\end{equation}
\end{corollary}
\begin{remark}[Free from distribution shift issues]
\label{rem:dist}
Since \eqref{eq:estimator} holds for all $x$ where the mixture $p_x(x)>0$, the estimator $\hat{r}_{p/q}(x)$ in \eqref{eq:rhat-def} is valid for all $x$ in the union of the domain of $p, q, m_1, \ldots, m_K$. This means that \cob does not suffer from the distribution shift problems that occur when solving a sequence of binomial logistic regression problems as in TRE. We exemplify this in Section~\ref{sec:distributionshift} after introducing three schemes to construct the auxiliary distributions $m_1, \ldots, m_K$. 
% We will expand on this in the next section.
\end{remark}

\subsection{Constructing the auxiliary distributions}
\label{sec:constructing_m}
In \cob, auxiliary distributions need to be constructed such that they have overlapping support with the empirical densities of $p$ and $q$. This allows the multi-class classification probabilities to be better calibrated and leads to an accurate density ratio estimation. We demonstrate this in panel (c) of Figure~\ref{fig:boundary_and_log_ratio}, where $p=\mathcal{N}(-1,0.1)$ and $q=\mathcal{N}(1,0.2)$ and the single auxiliary distribution $m$ is set to be Cauchy $\mathcal{C}(0,1)$ that clearly overlaps with the other two distributions. The classification probabilities are shown as the scatter plot that is overlayed on the empirical densities of these distributions. Compared to the BDRE case in panel (a), which has high confidence in regions without any data, the multi-class classifier assigns, for $p$ and $q$, high class probabilities only over the support of the data and not where there are barely any data points from these two distributions. Moreover, the auxiliary distribution well covers the space where $p$ and $q$ have low density, which provides the necessary training data to inform the values of $\hat{h}_1(x)$ and $\hat{h}_2(x)$ in that area, which leads to an accurate estimate of the log-density ratio shown in panel (d). This is contrast to BDRE in panel (a) where the classifier, while constrained enough to get the classification right, is not learned well enough to also get the density ratio right (panel b). This subtle, yet important distinction between the usage of auxiliary distributions in \cob compared to BDRE and TRE enables \cob to generalize on out-of-domain samples, as we will demonstrate in Section~\ref{sec:1dexp}. 

Next, we briefly describe three schemes to construct auxiliary distributions for \cob and leave the details to Appendix \ref{app:constructing_m}:
\textbf{(1) Overlapping Distribution }
Unlike TRE, the formulation of \cob does not require ``gradually bridging'' the two distributions $p$ and $q$, hence, we introduce a novel approach to constructing auxiliary distributions. We define $m_k$ as any distribution whose samples overlap with both $p$ and $q$, and $p << m_k$, $q << m_k$. This includes heavy-tailed distributions (e.g. Cauchy, Student-t), normal distributions, uniform distributions, or their mixtures. We use this scheme in all low-dimensional simulations.
\textbf{(2) Linear Mixing } In this scheme, $m_k$ is defined as the distribution of the samples generated by linearly combining samples $X_p = \{x_p^i\}_{i=1}^N$ and $X_q = \{x_q^i\}_{i=1}^N$ from distributions $p$ and $q$, respectively. The generative process for a single sample $x_{m_k}^i$ from $m_k$ is given by  $x_{m_k}^i = (1-\alpha_k) x_p^i + \alpha_k x_q^i$, with $x_p \in X_p, x_q^i \in X_q$. This construction is similar to the linear combination scheme for auxiliary distributions introduced by \citet{tre} with a few key differences that we expand upon in Appendix \ref{app:constructing_m}. One difference is that  $\alpha_k$ is not limited t
o $0 < \alpha_k < 1$, which allows for non-convex mixtures that completely surround both $p$ and $q$. We use this construction scheme in higher-dimensional simulations.
\textbf{(3) Dimension-wise Mixing } This construction scheme was introduced in \citep{tre}. Samples from the single auxiliary distribution $m$ are obtained by combining different subsets of dimensions from samples from $p$ and $q$. We use this scheme for experiments involving high-dimensional image data.

\subsection{Free from distribution-shift problems}
\label{sec:distributionshift}
{We continue with the example task of estimating the density ratio
between p = N (-1, 0.1) and q = N (1, 0.2) and here illustrate Remark
3.5 that \cob does not suffer from the distribution shift problem
identified in Section \ref{sec:dist_shift-tre}.}
% We continue with the example task of estimating the density ratio between $p=\mathcal{N}(-1,0.1)$ and $q=\mathcal{N}(1,0.2)$ and demonstrate that \cob does not suffer from the distribution shift problem identified in Section \ref{sec:dist_shift-tre}. 
We test \cob with two types auxiliary distributions. First using a heavy-tailed distribution ($m=\cauchydist(0,1)$), under the overlapping distributions scheme, and second, with waymark distributions $m_1, m_2, m_3$ as used by TRE in Figure~\ref{fig:dist_shift-tre} using their linear-mixing construction scheme.
% this in Figure \ref{fig:1dcobdemo}.
\begin{figure}[t]
    \centering
    \includegraphics[width=\textwidth]{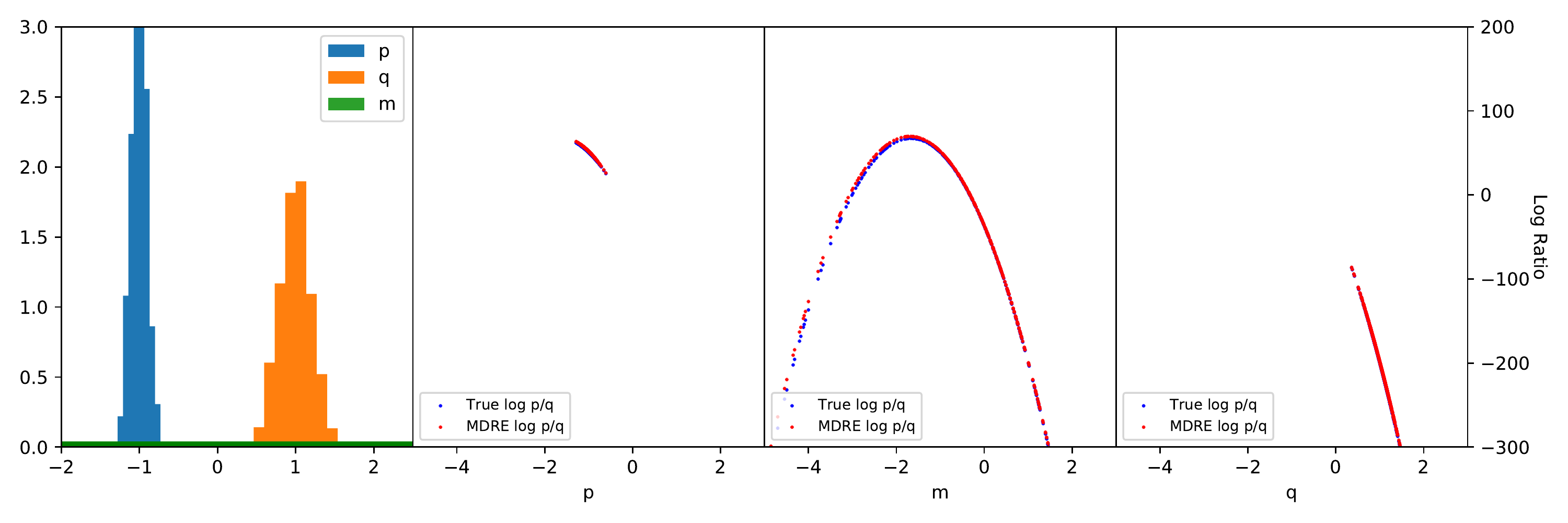}
    \caption{\cob with $m=\cauchydist(0,1)$. The first density plots shows the target densities as well as the auxiliary distribution (in green, hard to see due to heavy tail and the range of axes). The three scatter plots show the estimated (red) and true (blue) density ratio $p/q$ evaluated on samples from $p,m$, and $q$. \cob accurately estimates the ratio across the input domain. Contrast with Figure \ref{fig:dist_shift-tre}.}
    \label{fig:1dcobdemo}
\end{figure}

\begin{figure}[t]
    \centering
    \includegraphics[width=\textwidth]{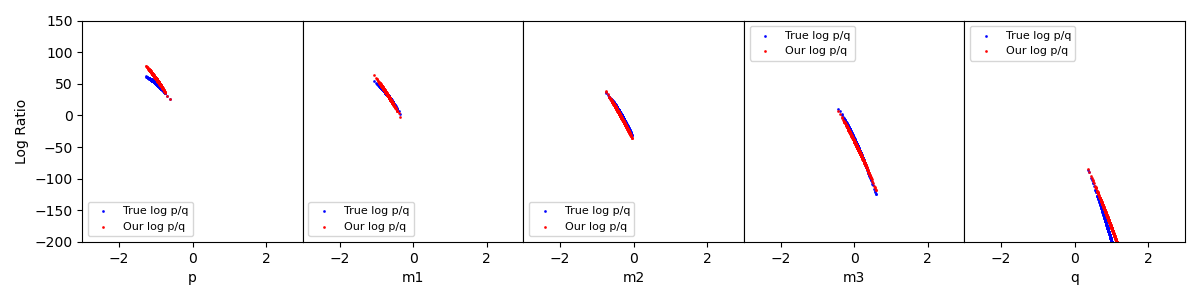}
    \caption{\cob using TRE's auxiliary distributions. Each scatter plot shows the overall log-density ratio estimates on samples from the distribution on the x-axis (\cob in red and true ratio in blue). \cob is capable of accurately estimating ratios on all samples. Contrast with the bottom row of Figure \ref{fig:dist_shift-tre}.}
    \label{fig:cob_insight_2}
\end{figure}

Figure~\ref{fig:1dcobdemo} shows the result for the heavy tailed $\cauchydist(0,1)$ auxiliary distribution (green, shown in the left most figure). 
We can see that the log-ratio learned by \cob is accurate even beyond the empirical support of $p$ and $q$. This is because \cob is trained on samples from the mixture of $p, q$ and $m$ and hence, per Remark \ref{rem:dist}, does not encounter distribution-shift, over the support of the mixture distribution.
Figure~\ref{fig:cob_insight_2} shows the result when using the auxiliary distributions of TRE that we used in Figure~\ref{fig:dist_shift-tre}. We see that the learned log-ratio well matches the true log-ratio on samples from $p$ and $q$, as well as the auxiliary distributions. This can be directly compared to third row of Figure~\ref{fig:dist_shift-tre} where TRE suffers from distribution shift problems and does not yield a well-estimated log-ratio. Note that we do not present results that correspond to the second row of \ref{fig:dist_shift-tre} since the estimation of the log-ratio in \cob \emph{does not} depend on any intermediate density ratios.

% \paragraph{TRE vs \cob:}
% While TRE turns an easy binary classification problem into \emph{a sequence} of harder binary classification problems, \cob turns the same easy binary classification problem into \emph{a single} harder multi-class classification problem. The sequence of binary classification problems do not share the same training domain while the single multi-class classification problem only has one training domain while used for multiple purposes (different ratio estimators). Another way to view this is that, for TRE, the original problem is a composition of the new problems, while for \cob, the original problem is a “sub-problem” of a new one and thus avoids distribution shift.

% It is worth pointing out that while here we relate a multinomial logistic regression to DRE, other class-probability estimators, in the multi-class setting, may also be related to density ratio estimation \cite{nock2016scaled}.

\section{Experiments}
\label{sec:exp}
We here provide an empirical analysis of \cob on both synthetic and real data, showing that it performs better than previous methods---\bdre, TRE, and F-DRE---on three different density ratio estimation tasks. We consider cases where numerator and denominator densities differ because their mean is different, i.e.\ $p$ and $q$ exhibit first-order discrepancies (FOD), and cases where the difference stems from different higher-order moments (higher-order discrepancies, HOD). Ratio estimation is closely related to KL divergence and mutual information estimation since the KL divergence is the expectation of the log-ratios under $p$, and mutual information can be expressed as a KL divergence between joint and product of marginals. Being quantities of core interest in machine learning, we will use them to evaluate the ratio estimation methods. 
%First, we compare \cob to all baselines on toy 1D Gaussian datasets where the two distributions differ in their first-order and second order moments (i.e. their means and variance).  We refer to discrepancies between distributions due to the difference between their first-order moments as First Order Discrepancies (FOD) and discrepancies due to the difference in higher-order moments as Higher Order Discrepancies (HOD). Then, we compare \cob to all baselines on a number of high-dimensional datasets, showing significant improvements on multiple tasks ranging from mutual information (MI) estimation to representation learning.
% \looseness=-1 Finally, we also show how \cob can also be applied to improve model-free DREs.
\subsection{1D Gaussian experiments with large KL divergence}
\label{sec:1dexp}

\begin{table*}[]
\centering
\begin{tabular}{@{}lllllll@{}}
\toprule
$p$                     & $q$                    & True KL & BDRE           & TRE                & F-DRE            & \cob (ours)              \\ \midrule
% $\mathcal{N}(-10, 1)$ & $\mathcal{N}(10, 1)$ & 200  & 34.45 $\pm$ 9.12 & 175.95 $\pm$ 1.88 & 18.41 $\pm$ 3.54 & 207.99 $\pm$ 2.30 \\
$\mathcal{N}(-1, 0.08)$ & $\mathcal{N}(2, 0.15)$ & 200.27  & 21.74 $\pm$ 4.10 & 136.05 $\pm$ 5.91  & 14.87 $\pm$ 1.72 & 203.32 $\pm$ 2.01 \\
$\mathcal{N}(-2, 0.08)$ & $\mathcal{N}(2, 0.15)$ & 355.82  & 20.22 $\pm$ 3.64 & 208.11 $\pm$ 18.31 & 14.22 $\pm$ 5.30 & 360.35 $\pm$ 1.37 \\ \hline
\end{tabular}
\caption[]
{\small 1D density ratio estimation task for $p$ and $q$ with large first-order and higher-order differences. In all cases, \cob outperforms all the baselines.}
\label{1dtask}
\end{table*}
% \begin{table*}[ht!]
% \centering
% \begin{tabular}{@{}lllllll@{}}
% \toprule
% $p$            & $q$      &     GT-KL & BC-DRE & TRE & F-DRE & \cob \\ \midrule
% $\mathcal{N}(0, \num{1e-6})$     & $\mathcal{N}(0, 1)$    & 13.32          & 3.06 $\pm$ 0.13                   & 8.12e-5 $\pm$ 4.52e-5        & 4.15 $\pm$ 0.31           & 13.14 $\pm$ 0.50                        \\
% $\mathcal{N}(-10, 1)$    & $\mathcal{N}(10, 1)$    & 200             & 34.45 $\pm$ 9.12                &   175.95 $\pm$ 1.88      &     18.41 $\pm$ 3.54     & 207.99 $\pm$ 2.30                        \\
% $\mathcal{N}(-1, 0.08)$ & $\mathcal{N}(2, 0.15)$ & 200.27         & 21.74 $\pm$ 4.10                 & 136.05 $\pm$ 5.91      & 14.87 $\pm$ 1.72         & 203.32 $\pm$ 2.01                       \\
% $\mathcal{N}(-2, 0.08)$ & $\mathcal{N}(2, 0.15)$ & 355.82         & 20.22 $\pm$ 3.64                  & 208.11 $\pm$ 18.31       & 14.22 $\pm$ 5.30            & 360.35 $\pm$ 1.37                      \\ \bottomrule
% \end{tabular}
% \caption[]
% {\small 1D density ratio estimation task. GT-KL stands for ground-truth KL Divergence.} 
% \label{1dtask}
% \end{table*}
In the following 1D experiments, we consider two scenarios, one where $p=\mathcal{N}(-1, 0.08)$ and $q=\mathcal{N}(2, 0.15)$, and one where the mean of $p$ is shifted to $-2$ in order to increase the degree of separation between the two distributions. In both cases, \cob's auxiliary distribution $m$ is $\mathcal{C}$auchy(0,1), so that we have a three-class classification problem ($p$, $q$, $m$) and three functions $h_\theta^i$ that parameterize the classifier of \cob. The three functions are quadratic polynomials of the form $w_1x^2+w_2x+b$. For all the methods we set the total number of samples to 100K.\footnote{We found that \cob's results are unchanged even when using smaller sample sizes of 1K or 10K, see Table \ref{tab:1dtask_app} in Appendix~\ref{app:1d_viz}.} We provide the exact hyperparameter settings for \cob and other baselines in Table \ref{tab:app_1dtask_configs} in Appendix \ref{app:1d_viz}.

\begin{figure*}[]
    \centering
    \begin{subfigure}[b]{0.49\textwidth}
        \includegraphics[width=\textwidth]{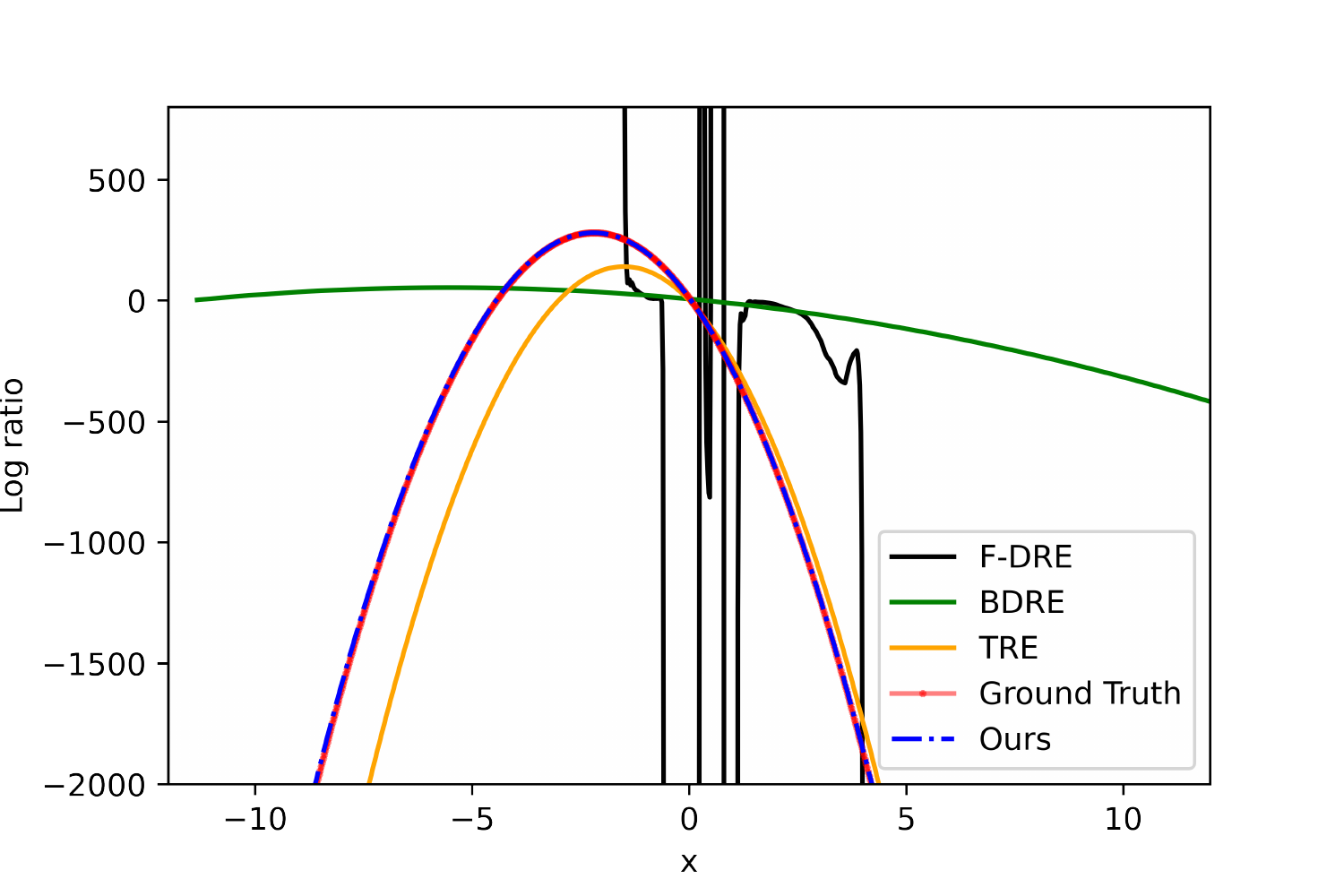}
        \caption{$p=\mathcal{N}(-1, 0.08), q=\mathcal{N}(2, 0.15)$}
        \label{fig:1d_ratios_-1_2_inoutdom}
    \end{subfigure}
    \begin{subfigure}[b]{0.49\textwidth}
        \includegraphics[width=\textwidth]{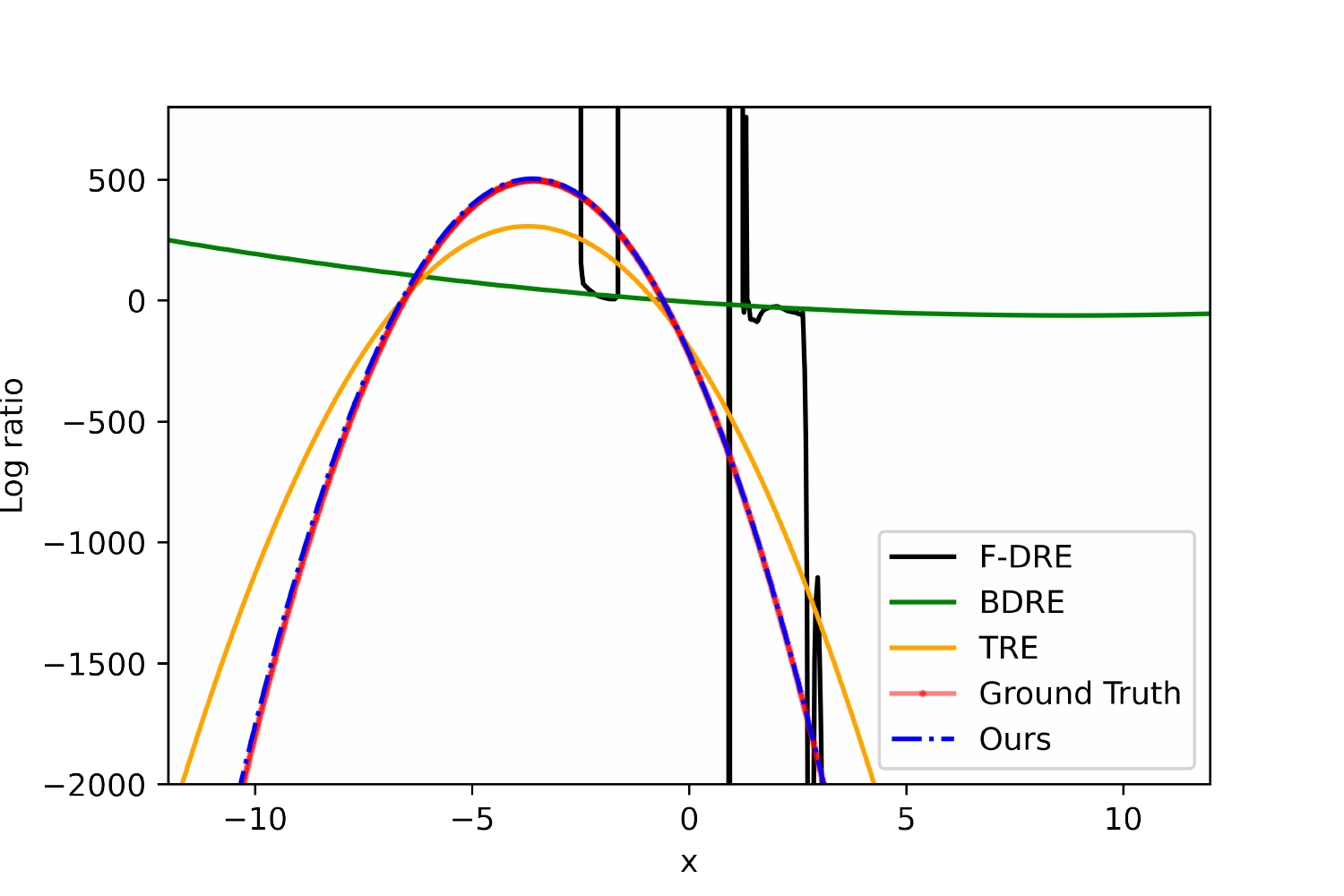}
        \caption{$p=\mathcal{N}(-2, 0.08), q=\mathcal{N}(2, 0.15)$}
        \label{fig:1d_ratios_-2_2_inoutdom}
    \end{subfigure}
    \caption{\small Log density-ratio estimates corresponding to the numbers reported in Table \ref{1dtask}. Note that the ground truth and \cob curves are overlapping, while all the other estimators are significantly worse.} \label{fig:1d_ratios_plot} 
\end{figure*}

Table \ref{1dtask} shows the results. We can see that \cob yields more accurate estimates of the KL divergences than the baselines, which are off by a significant margin.

We note that KL estimation only requires evaluating the log-ratios on samples from the numerator distribution $p$. In Figure \ref{fig:1d_ratios_plot}, we thus show results for all methods where we evaluate the estimated log-ratios on a wide interval (-12, 12). The figure shows that none of the baseline methods can accurately estimate the ratio well on the whole interval while \cob performs well overall. This is important because it means that the ratio is well estimated in regions where $p$ and $q$ have little probability mass. %In other words, the ratio estimated by \cob performs well when evaluated on out-of-training-distribution samples.
These results demonstrate the effectiveness of \cob with a single auxiliary distribution whose samples overlaps with those from both $p$ and $q$, in lieu of using a chain of BDREs with up to $K=28$ closely-packed auxiliary distributions as used by TRE. Please see Appendix \ref{app:1d_viz} for additional results and details.
\begin{wrapfigure}{r}{0.45\textwidth}
    \centering
    % <left> <lower> <right> <upper>  
    \includegraphics[width=.4\textwidth, trim = {1cm .5cm 1cm 0.18cm} ]{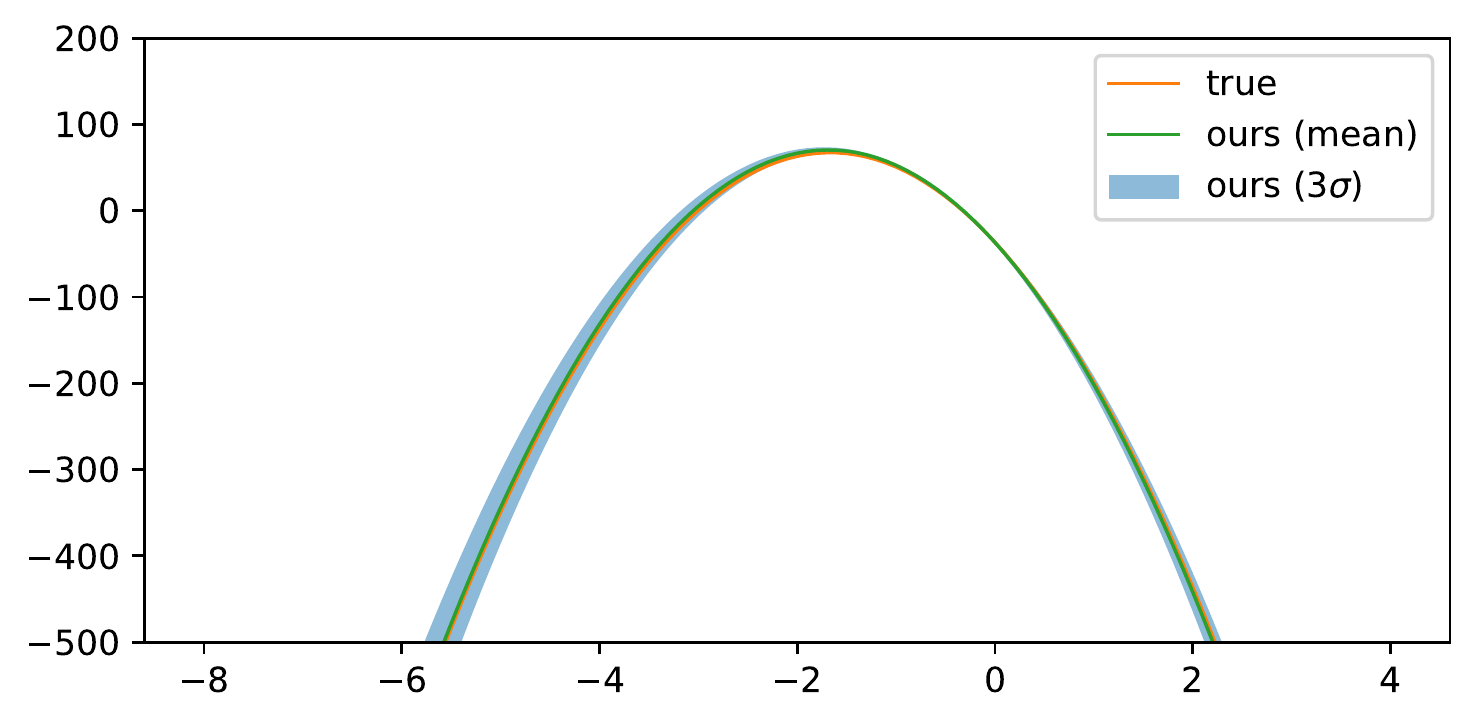}
    %\caption{{\small Full Bayesian analysis of \cob estimated using Hamiltonian Monte-Carlo: Mean (green) and $3\times$ standard deviation (light-blue) of $\log\frac{p}{q}$ for $p=\mathcal{N}(-1.0,0.1), q=\mathcal{N}(1.0,0.2)$. 
    %\cob has low uncertainty and higher accuracy in regions of high probability under $p$ and %$q$.}}\label{fig:ratio_hmc_intext}
    \caption{{\small Bayesian analysis of \cob for $p=\mathcal{N}(-1.0,0.1), q=\mathcal{N}(1.0,0.2)$. }}\label{fig:ratio_hmc_intext}
\end{wrapfigure}

To provide further clarity into \cob's density ratio estimation behavior, we analyze the uncertainty of its log ratio estimates using Bayesian analysis. We use a standard normal prior on the classifier parameters and obtain posterior samples with Hamiltonian Monte-Carlo. These posterior samples then yield samples of the density ratio estimates. Figure \ref{fig:ratio_hmc_intext} shows that the high accuracy of \cob's KL divergence estimates can be attributed to \cob being confidently accurate around the union of the high density regions of both $p$ and $q$. 
A more detailed analysis is provided in Appendix \ref{app:hmc}.
%\looseness=-1

\subsection{High dimensional experiments with large MI}
\label{sec:high_dim}
Following \citet{tre}, we use the MI estimation benchmark from \citet{belghazi2018mine, poole2019variational} to evaluate \cob on a more challenging, higher-dimensional problem. In this task, the goal is to estimate the mutual information between a standard normal distribution and a Gaussian random variable $x \in \mathbb{R}^{2d}$ with a block-diagonal covariance matrix where each block is $2 \times 2$ with ones on the diagonal and $\rho$ on the off-diagonal. The correlation coefficient $\rho$ is computed from the number of dimensions and the target mutual information $I = -d/2\log(1-\rho^2)$. Since this problem construction only induces higher-order discrepancies (HOD), we added an additional challenge by moving the means of the two distributions, thus additionally inducing first-order discrepancies (FOD).

For \cob, we model the $h_\theta^i$ with quadratic functions of the form $x^TW_1x+W_2x+b$. We use linear-mixing to construct each $m_k$, where $K=3$ or $K=5$. In Appendix \ref{app:high_dim}, we provide the exact configurations for \cob in Table \ref{tab:apphighdimconfig} and explain how to choose $m$ and $K$ in practice.  

\begin{table*}[]
\centering
\begin{tabular}{@{}cllllll@{}}
\toprule
\multicolumn{1}{l}{Dim} & $\mu_1, \mu_2$ & \multicolumn{1}{l}{True MI} & \multicolumn{1}{l}{BDRE} & \multicolumn{1}{l}{TRE} & F-DRE & \cob (ours)\\ \midrule
\multirow{2}{*}{40}  & 0, 0      & 20  & 10.90 $\pm$ 0.04 & 14.52 $\pm$ 2.07  & 14.87 $\pm$ 0.33 & 18.81 $\pm$ 0.15  \\
                     & -1, 1     & 100 & 29.03 $\pm$ 0.09 & 33.95 $\pm$ 0.14  & 13.86 $\pm$ 0.26 & 119.96 $\pm$ 0.94 \\ \midrule
\multirow{2}{*}{160} & 0, 0      & 40  & 21.47 $\pm$ 2.62 & 34.09 $\pm$ 0.21  & 12.89 $\pm$ 0.87 & 38.71 $\pm$ 0.73  \\
                     & -0.5, 0.6 & 136 & 24.88 $\pm$ 8.93 & 69.27 $\pm$ 0.24  & 13.74 $\pm$ 0.13 & 133.64 $\pm$ 3.70 \\ \midrule
\multirow{2}{*}{320} & 0, 0      & 80  & 23.47 $\pm$ 9.64 & 72.85 $\pm$ 3.93  & 9.17 $\pm$ 0.60  & 87.76 $\pm$ 0.77  \\
                     & -0.5, 0.5 & 240 & 24.86 $\pm$ 4.07 & 100.18 $\pm$ 0.29 & 10.53 $\pm$ 0.03 & 217.14 $\pm$ 6.02 \\ \bottomrule
\end{tabular}
\caption[ht!]
{\small High-dimensional mutual information estimation task. \cob is able to accurately estimate the MI often by a very large margins.} 
\label{tab:highdim}
\end{table*}

Table \ref{tab:highdim} shows the results for each MI task averaged across 3 runs with different random seeds. \cob outperforms all baselines in the original MI task where the means of the distribution are the same. The difference between the performance of \cob and the baselines is particularly stark when the means are allowed to be nonzero. Only \cob estimates the MI reasonably well while all baselines dramatically underestimate it. %This further demonstrates how \cob solves distribution shift when estimating the log ratio of the two densities. 
We further note that  \cob only uses up to $5$ auxiliary distributions, lowering its compute requirements compared to TRE, which is the next best performing method and uses up to $15$ auxiliary distributions for its telescoping chain. 

We found that the resolution proposed by \cite{kato2021non} to overcome the over-fitting issue in Bregman Divergence minimization-based DREs, does not work well in practice. On the high-dimensional setup of row 2 in Table \ref{tab:highdim}, while the ground truth MI is 100, and \cob estimates it as $119\pm0,94$, the best model from \citet{kato2021non} yields $1.60$, significantly underestimating the true value and being a factor of ten smaller than the classifier-based DRE baselines. For further results, such as plots of estimated log ratio vs. ground-truth log ratio, training curves, and more, please see Appendix \ref{app:high_dim}.

Above, following prior work, we evaluated the methods on problems where $p$ and $q$ are normal distributions. To enhance this analysis, we further evaluate \cob on the three new experimental setups below. The results are summarized in Table \ref{tab:failure}.

% \paragraph{Robustness and Generalization beyond Gaussian Distributions:} 
%\subsection{Robustness and Generalization beyond Gaussian Distributions}

\textbf{Breaking Symmetry}
In our high-dimensional experiments reported in Table \ref{tab:highdim}, the means of the Gaussian distributions $p$ and $q$ were symmetric around zero in the majority of cases. In order to ensure that this symmetry did not provide an advantage to \cob, we also evaluate it on Gaussians $p$ and $q$ with randomized means. The results are shown in rows 2 and 6 of Table \ref{tab:failure}. We see that \cob continues to estimate the ground truth KL divergence accurately, demonstrating that it did not benefit unfairly from the symmetry of distributions around zero.\looseness=-1

\textbf{Model Mismatch}
In rows 4, 5, 7, and 8 of Table \ref{tab:failure}, we evaluate \cob by replacing one or both distributions $p$ and $q$ with a Student-t distribution of the same scale with randomized means. For the Student-t distributions, we set the degrees of freedom as $5$, $10$ or $20$. These experiments test how well \cob performs when there is model mismatch, i.e. how \cob performs using the same quadratic model that was used when $p$ and $q$ were set to be Gaussian with lighter tails.
We find that \cob is still able to accurately estimate the ground truth KL in these cases. We found the same to be true for other test distributions such as a Mixture of Gaussians (shown in row 3 of Table \ref{tab:failure}).

\textbf{Finite Support $p$ and $q$} 
Finally, we test \cob on another problem where $p$ and $q$ are finite support distributions that have both FOD and HOD. This is done by setting $p$ and $q$ to be truncated normal distributions, as shown in row 1 of Table \ref{tab:failure}. We also set $m$ to be a truncated normal distribution with its scale set to 2 to allow it to have overlap with both $p$ and $q$. This setting is similar to the 1D Gaussian example illustrated in Section \ref{sec:distributionshift} and \cob manages to estimate the ground-truth KL divergence accurately. 
% Figure \ref{fig:truncated} shows that \cob manages to well estimate the ground truth KL, and, furthermore, to correctly decay its log ratio estimates towards negative infinity when the samples do not fall under the support of $p$.

\begin{table*}[ht!]
\centering
\begin{tabular}{@{}llllll@{}}
\toprule
Dim &
  $p$ &
  $q$ &
  $m$ &
  True KL &
  Est. KL \\ \midrule
1 &
  \begin{tabular}[c]{@{}l@{}}Truncated Normal\\  loc=-1, scale=0.1 \\ support=(-1.1,-0.9)\end{tabular} &
  \begin{tabular}[c]{@{}l@{}}Truncated Normal\\ loc=1, scale=0.2 \\ support=(-1.1,1.2)\end{tabular} &
  \begin{tabular}[c]{@{}l@{}}Truncated Normal \\ loc=-1, scale=2 \\ support=(-1.1,1.2)\end{tabular} &
  50.65 &
  52.35 \\ \midrule
160 &
  \begin{tabular}[c]{@{}l@{}}Normal \\ loc=R(-.5,.5), cov=$2\times2$ BD\end{tabular} &
  \begin{tabular}[c]{@{}l@{}}Normal \\ loc=R(-.5,.5), cov=$I$\end{tabular} &
  Linear Mixing &
  54.29 &
  54.10 \\ \midrule
160 &
  \begin{tabular}[c]{@{}l@{}}Normal \\ loc=-1, cov=$2\times2$ BD\end{tabular} &
  \begin{tabular}[c]{@{}l@{}} MoG: 0.5*Normal(0.9, $I$)\\  + 0.5*Normal(1.1,$I$)\end{tabular} &
  Linear Mixing &
  105.60 &
  98.27 \\ \midrule
160 &
  \begin{tabular}[c]{@{}l@{}}Student T loc=R(-.5,.5), \\ scale=$2\times2$ BD, df=5\end{tabular} &
  \begin{tabular}[c]{@{}l@{}}Student T \\ loc=R(-.5,.5), scale=I, df=5\end{tabular} &
  Linear Mixing &
  51.26 &
  49.01 \\ \midrule
320 &
  \begin{tabular}[c]{@{}l@{}}Student T loc=R(-.5,.5), \\ scale=$2\times2$ BD, df=10\end{tabular} &
  \begin{tabular}[c]{@{}l@{}}Student T \\ loc=R(-.5,.5), scale=I, df=10\end{tabular} &
  Linear Mixing &
  53.82 &
  51.03 \\ \midrule
320 &
  \begin{tabular}[c]{@{}l@{}}Normal \\ loc=R(-1,1), cov=$2\times2$ BD\end{tabular} &
  \begin{tabular}[c]{@{}l@{}}Normal \\ loc=R(-1,1), cov=$I$\end{tabular} &
  Linear Mixing &
  110.05 &
  102.63 \\ \midrule
320 &
  \begin{tabular}[c]{@{}l@{}}Student T loc=R(-1,1), \\ scale=$2\times2$ BD, df=10\end{tabular} &
  \begin{tabular}[c]{@{}l@{}}Student T \\ loc=R(-1,1), scale=$I$, df=10\end{tabular} &
  Linear Mixing &
  103.12 &
  113.53 \\ \midrule
320 &
  \begin{tabular}[c]{@{}l@{}}Normal \\ loc=0, cov=$2\times2$ BD\end{tabular} &
  \begin{tabular}[c]{@{}l@{}}Student T \\ loc=0, scale=$I$, df=20\end{tabular} &
  Linear Mixing &
  82.02 &
  83.63 \\ \bottomrule
\end{tabular}
\caption[]
{\small Robustness evaluation for \cob. Here R(a,b) stands for randomized mean vector where each dimension is sampled uniformly from the interval $(a,b)$. \cob is able to consistently estimate the ground-truth KL with high accuracy in all of the cases.} 
\label{tab:failure}
\vspace{-1em}
\end{table*}

\subsection{Representation learning for SpatialMultiOmniglot}
In order to benchmark \cob on large-scale real-world data, following the setup from \citet{tre}, we apply \cob to the task of mutual information estimation and representation learning for the SpatialMultiOmniglot problem \citep{ozair2019wasserstein}. The goal is to estimate the mutual information between $u$ and $v$ where $u$ is a $n \times n$ grid of Omniglot characters from different Omniglot alphabets and $v$ is a $n \times n$ grid containing (stochastic) realizations of the next characters of the corresponding characters in $u$. After learning, we evaluate the representations from the encoder with a standard linear evaluation protocol \citep{infonce}. For \cob, similarly to TRE, we utilize a separable architecture commonly used in the MI-based representation learning literature and model the unnormalized log-scores $h_\theta^i$ with functions of the form $g(u)^T Wf(v)$ where $g$ and $f$ are 14-layer convolutional ResNets \citep{he2015resnet}. {While this model amounts to sharing of parameters across the $h_\theta^i$, we would like to emphasize that in all preceding examples, we did not share parameters among the $h_\theta^i$.} We construct the auxiliary distributions via dimension-wise mixing. 

We here only compare \cob to the single ratio baseline and TRE because \citet[][Figure 4]{tre} already demonstrated that TRE significantly outperforms both Contrastive Predictive Coding (CPC) \citep{infonce} and Wasserstein Predictive Coding (WPC) \citep{ozair2019wasserstein} on exactly the same task. Please refer to Appendix \ref{app:omni} for the detailed experimental setup.

\begin{figure*}
    \centering
    \begin{subfigure}[b]{0.329\textwidth}
        \includegraphics[width=\textwidth]{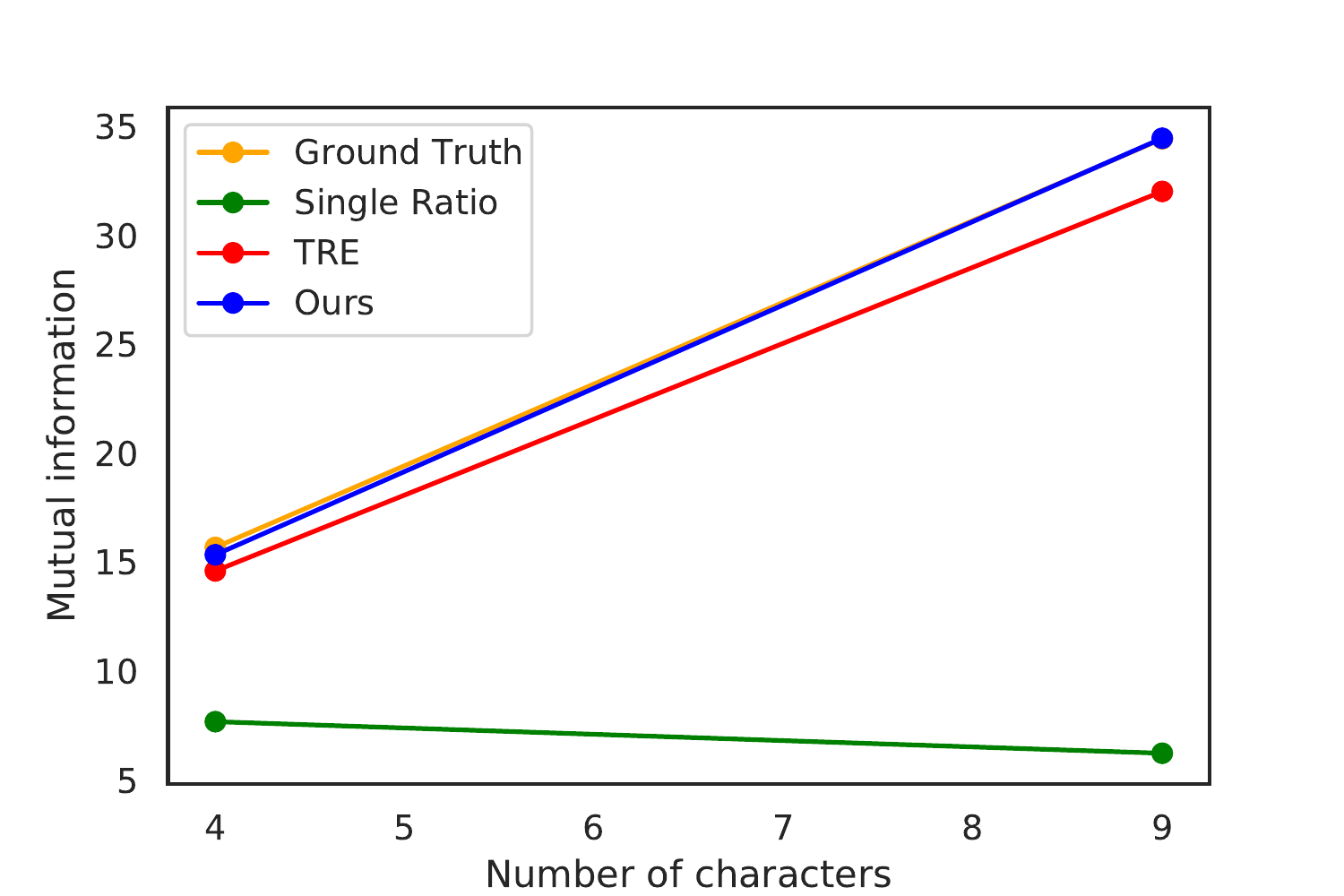}
        \caption{MI estimation}
        \label{fig:omni-mi}
    \end{subfigure}
    \begin{subfigure}[b]{0.329\textwidth}
        \includegraphics[width=\textwidth]{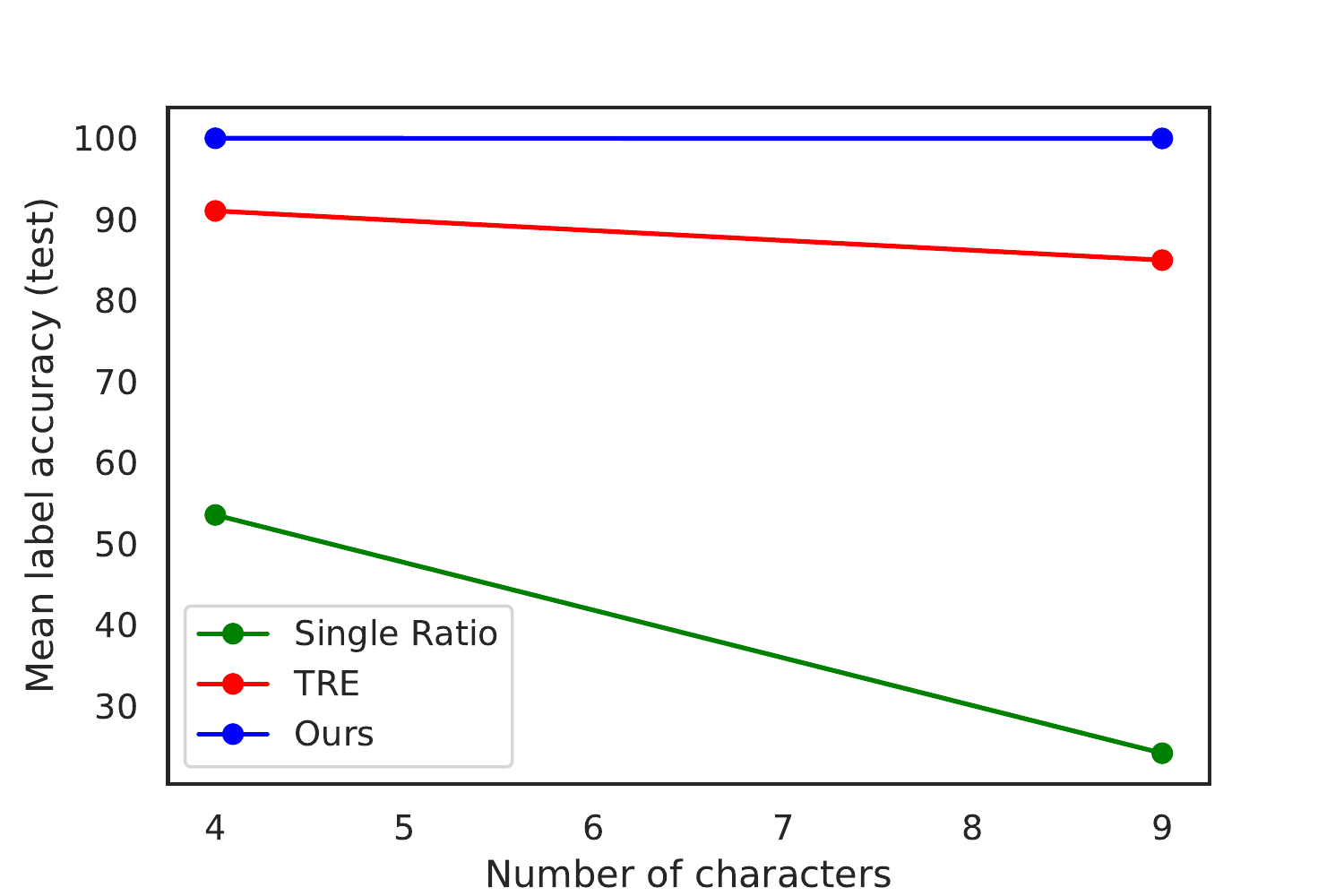}
        \caption{Classification accuracy}
        \label{fig:omni-acc}
    \end{subfigure}
    \begin{subfigure}[b]{0.329\textwidth}
        \includegraphics[width=\textwidth]{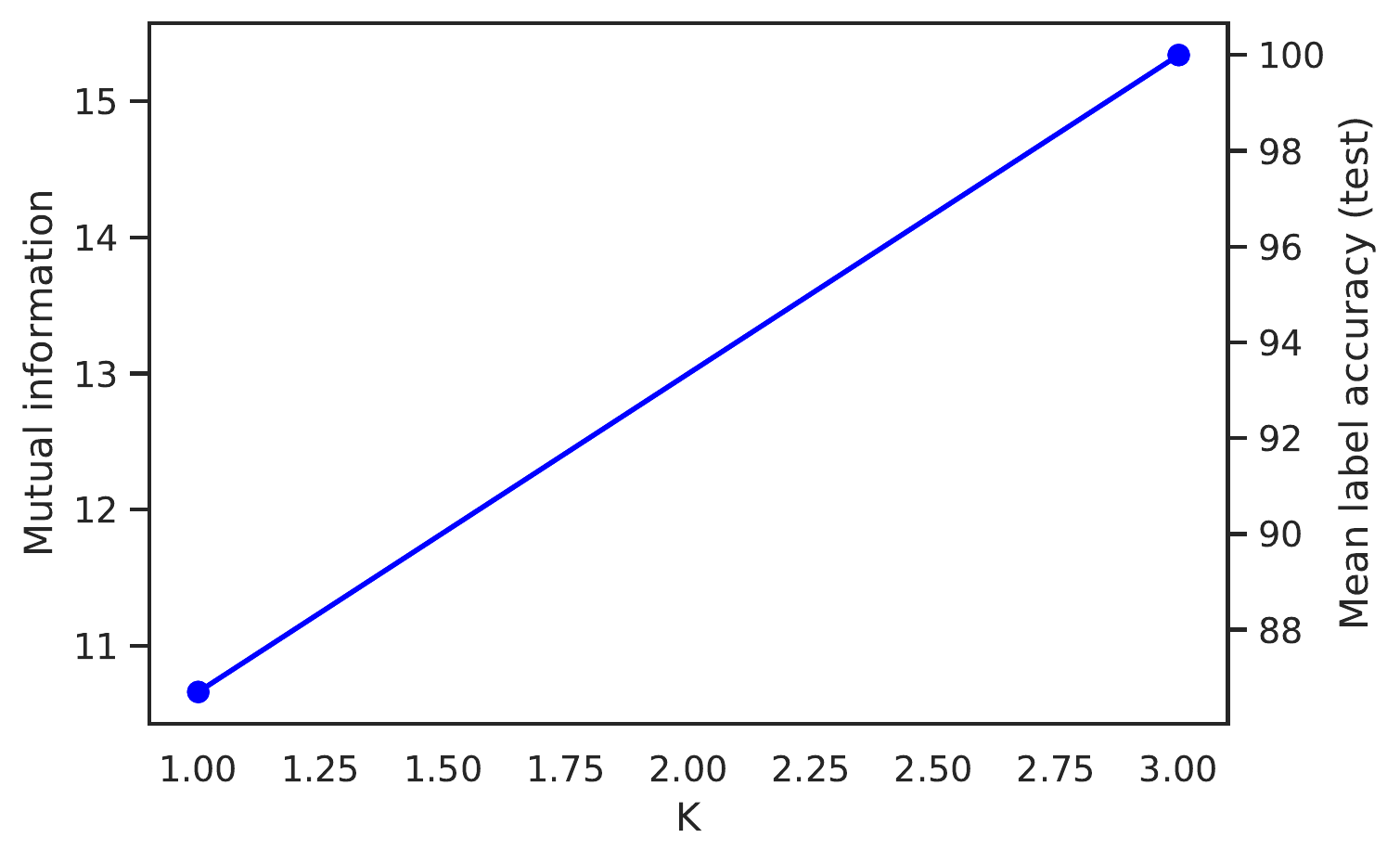}
        \caption{Varying $K$ (\# of auxiliary dist.)}
        \label{fig:omni_k}
    \end{subfigure}
    \caption{SpatialMultiOmniglot representation learning results. Plot (a) shows the MI estimated by the three methods, \cob is able to estimate the ground truth MI very accurately. Plot (b) shows the resulting classification accuracy and plot (c) the impact of varying the number of auxiliary distributions on MI estimation with \cob. }\label{fig:omni}
    \vspace{0em}
\end{figure*}
As can be seen in Figure \ref{fig:omni-mi}, \cob performs better than TRE and the single ratio baseline, exactly matching the ground truth MI. This improvement in MI estimation is reflected in the representations. Figure \ref{fig:omni-acc} illustrates that \cob's encoder learns representations that achieve $\sim$100\% Omniglot character classification for both $d=n^2=4, 9$. On the other hand, the performances of the single ratio estimator and TRE (using the same exact dimension-wise mixing to construct auxiliary distributions) both degrade as the complexity of the task increases, with TRE only reaching up to  91\% and 85\% for $d=4$ and $d=9$, respectively. All models were trained with the same encoder architecture to ensure fair comparison. 

We further studied the effect of changing $K$ in the $d=4$ setup. For $K=1$, we aggregate all the dimension-wise mixed samples into 1 class, whereas for $K=3$, we separate them into their respective classes (corresponding to the number of dimensions mixed). We illustrate this effect in Figure \ref{fig:omni_k}. In line with the finding of \citet{ma2018noise}, increasing the number of K not only helps \cob to reach the ground truth MI, but also the quality of representations improves from $86.7\%$ to $100\%$ test classification accuracy. 

\looseness=-1

\section{Discussion}
{In this work, we presented the multinomial logistic regression based
density ratio estimator (MDRE), a new method for density ratio
estimation that had better finite sample (non-asymptotic) performance in
our simulations than current state-of-the-art methods.}
% In this work, we presented the multinomial logistic regression based density ratio estimator (\cob), a new method for density ratio estimation that has better finite sample (non-asymptotic) behavior than current state-of-the-art methods. 
We showed that it addresses the sensitivity to possible distribution-shift issues of the recent method by \citet{tre}. \cob works by introducing auxiliary distributions that have overlapping support with the numerator and denominator distributions of the ratio. It then trains a multinomial logistic regression model to estimate the density-ratio.
We demonstrated that \cob is both theoretically grounded and empirically strong, and that it sets a new state of the art for high-dimensional density ratio estimation problems.

However, there are some limitations. First, while the ratio was well estimated in our empirical studies, we do not provide any bounds on the estimation, meaning that estimated KL divergences or mutual information values may be over- or underestimated. {Second, the choice of the auxiliary distribution, $m$, is an important factor of consideration that significantly impacts the performance of \cob. While in this work we demonstrate the efficacy of three schemes for constructing the auxiliary distribution, empirically, it is, by no means, an exhaustive study. We hope to address these issues in future work, including the development of learning-based approaches to auxiliary distribution construction. } 

\bibliography{iclr2021_conference}

\begin{thebibliography}{27}
\providecommand{\natexlab}[1]{#1}
\providecommand{\url}[1]{\texttt{#1}}
\expandafter\ifx\csname urlstyle\endcsname\relax
  \providecommand{\doi}[1]{doi: #1}\else
  \providecommand{\doi}{doi: \begingroup \urlstyle{rm}\Url}\fi

\bibitem[Belghazi et~al.(2018)Belghazi, Baratin, Rajeswar, Ozair, Bengio,
  Courville, and Hjelm]{belghazi2018mine}
Mohamed~Ishmael Belghazi, Aristide Baratin, Sai Rajeswar, Sherjil Ozair, Yoshua
  Bengio, Aaron Courville, and R~Devon Hjelm.
\newblock Mine: mutual information neural estimation.
\newblock \emph{arXiv preprint arXiv:1801.04062}, 2018.

\bibitem[Bickel et~al.(2008)Bickel, Bogojeska, Lengauer, and
  Scheffer]{bickel2008multi}
Steffen Bickel, Jasmina Bogojeska, Thomas Lengauer, and Tobias Scheffer.
\newblock Multi-task learning for hiv therapy screening.
\newblock In \emph{Proceedings of the 25th international conference on Machine
  learning}, pp.\  56--63, 2008.

\bibitem[Choi et~al.(2021{\natexlab{a}})Choi, Liao, and
  Ermon]{choi2021featurized}
Kristy Choi, Madeline Liao, and Stefano Ermon.
\newblock Featurized density ratio estimation.
\newblock \emph{arXiv preprint arXiv:2107.02212}, 2021{\natexlab{a}}.

\bibitem[Choi et~al.(2021{\natexlab{b}})Choi, Meng, Song, and
  Ermon]{choi2021density}
Kristy Choi, Chenlin Meng, Yang Song, and Stefano Ermon.
\newblock Density ratio estimation via infinitesimal classification.
\newblock \emph{arXiv preprint arXiv:2111.11010}, 2021{\natexlab{b}}.

\bibitem[Csisz{\'a}r(1964)]{csiszar1964}
Imre Csisz{\'a}r.
\newblock An information-theoretic inequality and its application to the
  evidence of the ergodicity of markoff's chains.
\newblock \emph{Magyer Tud. Akad. Mat. Kutato Int. Koezl.}, 8:\penalty0
  85--108, 1964.

\bibitem[Goodfellow et~al.(2014)Goodfellow, Pouget{-}Abadie, Mirza, Xu,
  Warde{-}Farley, Ozair, Courville, and Bengio]{gan}
Ian~J. Goodfellow, Jean Pouget{-}Abadie, Mehdi Mirza, Bing Xu, David
  Warde{-}Farley, Sherjil Ozair, Aaron~C. Courville, and Yoshua Bengio.
\newblock Generative adversarial nets.
\newblock In \emph{Neural Information Processing Systems}, 2014.

\bibitem[Gutmann \& Hirayama(2011)Gutmann and Hirayama]{Gutmann2011b}
M.~U. Gutmann and J.~Hirayama.
\newblock {B}regman divergence as general framework to estimate unnormalized
  statistical models.
\newblock In \emph{Proceedings of the Conference on Uncertainty in Artificial
  Intelligence (UAI)}, 2011.

\bibitem[Gutmann \& Hyv{\"a}rinen(2010)Gutmann and Hyv{\"a}rinen]{nce}
Michael Gutmann and Aapo Hyv{\"a}rinen.
\newblock Noise-contrastive estimation: A new estimation principle for
  unnormalized statistical models.
\newblock In \emph{Proceedings of the thirteenth international conference on
  artificial intelligence and statistics}, pp.\  297--304. JMLR Workshop and
  Conference Proceedings, 2010.

\bibitem[Gutmann \& Hyv\"arinen(2012)Gutmann and Hyv\"arinen]{Gutmann2012a}
M.U. Gutmann and A.~Hyv\"arinen.
\newblock {N}oise-contrastive estimation of unnormalized statistical models,
  with applications to natural image statistics.
\newblock \emph{Journal of Machine Learning Research}, 13:\penalty0 307--361,
  2012.

\bibitem[He et~al.(2015)He, Zhang, Ren, and Sun]{he2015resnet}
Kaiming He, Xiangyu Zhang, Shaoqing Ren, and Jian Sun.
\newblock Deep residual learning for image recognition, 2015.

\bibitem[Kato \& Teshima(2021)Kato and Teshima]{kato2021non}
Masahiro Kato and Takeshi Teshima.
\newblock Non-negative bregman divergence minimization for deep direct density
  ratio estimation.
\newblock In \emph{International Conference on Machine Learning}, pp.\
  5320--5333. PMLR, 2021.

\bibitem[Liu et~al.(2021)Liu, Rosenfeld, Ravikumar, and
  Risteski]{liu2021analyzing}
Bingbin Liu, Elan Rosenfeld, Pradeep Ravikumar, and Andrej Risteski.
\newblock Analyzing and improving the optimization landscape of
  noise-contrastive estimation.
\newblock \emph{arXiv preprint arXiv:2110.11271}, 2021.

\bibitem[Ma \& Collins(2018)Ma and Collins]{ma2018noise}
Zhuang Ma and Michael Collins.
\newblock Noise contrastive estimation and negative sampling for conditional
  models: Consistency and statistical efficiency.
\newblock \emph{arXiv preprint arXiv:1809.01812}, 2018.

\bibitem[Menon \& Ong(2016)Menon and Ong]{menon2016linking}
Aditya Menon and Cheng~Soon Ong.
\newblock Linking losses for density ratio and class-probability estimation.
\newblock In \emph{International Conference on Machine Learning}, pp.\
  304--313. PMLR, 2016.

\bibitem[Nock et~al.(2016)Nock, Menon, and Ong]{nock2016scaled}
Richard Nock, Aditya Menon, and Cheng~Soon Ong.
\newblock A scaled bregman theorem with applications.
\newblock \emph{Advances in Neural Information Processing Systems},
  29:\penalty0 19--27, 2016.

\bibitem[Nowozin et~al.(2016)Nowozin, Cseke, and Tomioka]{fgan}
Sebastian Nowozin, Botond Cseke, and Ryota Tomioka.
\newblock f-gan: Training generative neural samplers using variational
  divergence minimization.
\newblock In \emph{Proceedings of the 30th International Conference on Neural
  Information Processing Systems}, pp.\  271--279, 2016.

\bibitem[Oord et~al.(2018)Oord, Li, and Vinyals]{infonce}
Aaron van~den Oord, Yazhe Li, and Oriol Vinyals.
\newblock Representation learning with contrastive predictive coding.
\newblock \emph{arXiv preprint arXiv:1807.03748}, 2018.

\bibitem[Ozair et~al.(2019)Ozair, Lynch, Bengio, van~den Oord, Levine, and
  Sermanet]{ozair2019wasserstein}
Sherjil Ozair, Corey Lynch, Yoshua Bengio, Aaron van~den Oord, Sergey Levine,
  and Pierre Sermanet.
\newblock Wasserstein dependency measure for representation learning, 2019.

\bibitem[Poole et~al.(2019)Poole, Ozair, van~den Oord, Alemi, and
  Tucker]{poole2019variational}
Ben Poole, Sherjil Ozair, Aaron van~den Oord, Alexander~A. Alemi, and George
  Tucker.
\newblock On variational bounds of mutual information, 2019.

\bibitem[Qui{\~n}onero-Candela et~al.(2009)Qui{\~n}onero-Candela, Sugiyama,
  Lawrence, and Schwaighofer]{quinonero2009dataset}
Joaquin Qui{\~n}onero-Candela, Masashi Sugiyama, Neil~D Lawrence, and Anton
  Schwaighofer.
\newblock \emph{Dataset shift in machine learning}.
\newblock Mit Press, 2009.

\bibitem[Radford et~al.(2015)Radford, Metz, and Chintala]{dcgan}
Alec Radford, Luke Metz, and Soumith Chintala.
\newblock Unsupervised representation learning with deep convolutional
  generative adversarial networks.
\newblock \emph{arXiv preprint arXiv:1511.06434}, 2015.

\bibitem[Rezende \& Mohamed(2015)Rezende and Mohamed]{rezende2015variational}
Danilo Rezende and Shakir Mohamed.
\newblock Variational inference with normalizing flows.
\newblock In \emph{International conference on machine learning}, pp.\
  1530--1538. PMLR, 2015.

\bibitem[Rhodes et~al.(2020)Rhodes, Xu, and Gutmann]{tre}
Benjamin Rhodes, Kai Xu, and Michael~U Gutmann.
\newblock Telescoping density-ratio estimation.
\newblock \emph{arXiv preprint arXiv:2006.12204}, 2020.

\bibitem[Srivastava et~al.(2017)Srivastava, Valkov, Russell, Gutmann, and
  Sutton]{veegan}
Akash Srivastava, Lazar Valkov, Chris Russell, Michael~U Gutmann, and Charles
  Sutton.
\newblock Veegan: Reducing mode collapse in gans using implicit variational
  learning.
\newblock In \emph{Proceedings of the 31st International Conference on Neural
  Information Processing Systems}, pp.\  3310--3320, 2017.

\bibitem[Srivastava et~al.(2020)Srivastava, Xu, Gutmann, and Sutton]{gram}
Akash Srivastava, Kai Xu, Michael Gutmann, and Charles Sutton.
\newblock Generative ratio matching networks.
\newblock In \emph{Eighth International Conference on Learning
  Representations}, pp.\  1--18, 2020.

\bibitem[Sugiyama et~al.(2012)Sugiyama, Suzuki, and
  Kanamori]{sugiyama2012density}
Masashi Sugiyama, Taiji Suzuki, and Takafumi Kanamori.
\newblock \emph{Density ratio estimation in machine learning}.
\newblock Cambridge University Press, 2012.

\bibitem[Wasserman(2004)]{Wasserman2004}
L.~Wasserman.
\newblock \emph{{A}ll of statistics}.
\newblock Springer, 2004.

\end{thebibliography}

\bibliographystyle{tmlr}

\appendix
%\section{Appendix}
{
\section{Consistency}
\label{app:consistency}
 In Section \ref{sec:loss-function}, we focused on properties of the loss function $\mathcal{L}(h_1,\dots , h_C)$ in \eqref{eq:functional-multinomial-loss}. The arguments of the loss functions were the functions $h_i$, and the loss function was defined in terms of expectations over $p_c$. This simplified the analysis and provided important insights but does not correspond to practical settings. Here, we relax the assumptions: We first consider the loss $\mathcal{L}(\theta)$ in \eqref{eq:multinomial-loss} where the functions $h_c$ are parameterized by some parameters $\theta$. Then we consider the case where the expectations are replaced by a sample average based on $n$ samples. The corresponding loss function will be denote by $\mathcal{L}_n(\theta)$. The main point of this section is to derive conditions under which minimizing $\mathcal{L}_n(\theta)$ leads to the results in Section \ref{sec:loss-function}, obtained by minimizing $\mathcal{L}(h_1,\dots , h_C)$.

\begin{lemma}\label{le:minimiser-characterisation}
Denoting the true conditional distribution by $P^\ast(Y|x)$ we have
\begin{equation}
   \hat{\theta}= \argmin_\theta \mathcal{L}(\theta) = \argmin_\theta \mathbb{E}_{p_x(x)} \text{KL}\left(P^\ast(Y|x) \mid\mid P(Y|x; \theta)\right)
\end{equation}
where $P(Y|x; \theta)$ is defined in \eqref{eq:parametric-model}.
\end{lemma}
\begin{proof}
We start with the definition of $\mathcal{L}(\theta)$ in \eqref{eq:multinomial-loss}:
\begin{align}
\mathcal{L}(\theta)&= - \sum_{c=1}^C \pi_c \mathbb{E}_{x\sim p_c}[\log P(Y=c|x; \theta)]
\end{align}
The sum of weighted expectations $\sum_{c=1}^C \pi_c \mathbb{E}_{x\sim p_c}$ corresponds to a joint expectation over $p(Y,x)$. Decomposing the joint as $p(x)P^\ast(Y|x)$, we thus obtain
\begin{align}
    \mathcal{L}(\theta)&= - \mathbb{E}_{p(x)}\mathbb{E}_{P^\ast(Y|x)} [\log P(Y|x; \theta)]
\end{align}
and
\begin{align}
\mathcal{L}(\theta)+\mathbb{E}_{p(x)}\mathbb{E}_{P^\ast(Y|x)} \log P^\ast(Y|x) &= \mathbb{E}_{p(x)}\mathbb{E}_{P^\ast(Y|x)} \log \frac{P^\ast(Y|x)}{P(Y|x; \theta)}\\
& = \mathbb{E}_{p_x(x)} \text{KL}\left(P^\ast(Y|x) \mid\mid P(Y|x; \theta)\right)
\end{align}
The claim follows since the added term does not depend on $\theta$.
\end{proof}
If the true conditional $P^\ast(Y|x)$ is part of the parametric family $\{P(Y|x; \theta)\}_\theta$, $\hat{\theta}$ is thus such that $P(Y|x; \hat{\theta})=P^\ast(Y|x)$ for all $x$ where $p_x(x)>0$. Hence the same arguments after \eqref{eq:critical} in the main text lead the parametric equivalent to \eqref{prop:cob}, which we summarize in the following corollary.
\begin{corollary}\label{cor:estimator-parametric}
If the true conditional $P^\ast(Y|x)$ is part of the parametric family $\{P(Y|x; \theta)\}_\theta$, then
\begin{align}
    \label{eq:estimator-parametric}
    h^i_{\hat \theta}(x)- h^j_{\hat \theta}(x) = \log\frac{p_i(x)}{p_j(x)}
\end{align}
for all $x$ where $p_x(x)=\sum_c \pi_c p_c(x) >0$.
\end{corollary}
We next derive conditions under which $\hat \theta$ is the unique minimum, which is needed to prove consistency. For that purpose, we perform a second-order Taylor expansion of $\mathcal{L}(\theta)$ around $\hat{\theta}$.
\begin{lemma} \label{le:taylor-expansion}
\begin{align}
\mathcal{L}(\hat \theta + \epsilon \phi) &= \mathcal{L}(\theta) + \frac{\epsilon^2}{2} \phi^\top I \phi
\end{align}
where $\epsilon >0$ and $I = - \mathbb{E}_{p_x(x)}\mathbb{E}_{P^\ast(Y|x)}[H(Y,x)]$. The matrix $H(Y,x)$ contains the second derivatives of the log-model, i.e.\ its $(i,j)$-th element is
\begin{align}
    [H(Y,x)]_{ij} = \frac{\partial^2}{\partial \theta_i \partial \theta_j} \log P(Y|x; \theta)\bigg |_{\theta=\hat{\theta}}
\end{align}
where $\theta_i$ and $\theta_j$ are the $i$-th and $j$-th element of $\theta$, respectively.
\end{lemma}
\begin{proof}
A second-order Taylor expansion around $\mathcal{L}(\hat{\theta})$ gives
\begin{align}
    \mathcal{L}(\hat{\theta}+\epsilon \phi) =& - \mathbb{E}_{p_x(x)}\sum_c P^\ast(Y=c|x)\log P(Y=c|x, \hat{\theta}+\epsilon \phi) \\
    =& \mathcal{L}(\hat{\theta})- \nabla_\theta \mathcal{L}(\theta)\bigg |_{\theta=\hat{\theta}} -\mathbb{E}_{p_x(x)}\sum_c P^\ast(Y=c|x) \frac{\epsilon^2}{2}\phi^\top H(Y=c, x) \phi+O(\epsilon^2)\\
    =& \mathcal{L}(\hat{\theta}) -\frac{\epsilon^2}{2} \phi^\top \left[\mathbb{E}_{p_x(x)} \sum_c P^\ast(Y=c|x) H(Y=c, x)\right] \phi+O(\epsilon^2)
\end{align}
where we have used that the gradient of $\mathcal{L}(\theta)$ is zero at a minimizer $\hat{\theta}$. Since $\sum_c P^\ast(Y=c|x) H(Y=c, x)=\mathbb{E}_{P^\ast(Y|x)}H(Y, x)$, the result follows.
\end{proof}
Note that $I(x)=-\mathbb{E}_{P^\ast(Y|x)}[H(Y,x)]$ is the conditional Fisher information matrix, and $I = \mathbb{E}_{p_x(x)} I(x)$ is its expected value taken with respect to $p_x(x)$.
\begin{corollary}
If $I$ is positive definite, then $\hat \theta$ is the unique minimizer of $\mathcal{L}(\theta)$.
\label{cor:unique-minimizer}
\end{corollary}
\begin{proof}
If $I$ is positive definite, then $\phi^\top I \phi>0$ for all non-zero $\phi$ and by Lemma \ref{le:taylor-expansion}, $\mathcal{L}(\hat \theta + \epsilon \phi) > \mathcal{L}(\hat \theta)$ whenever $\phi \neq 0$.
\end{proof}
We now consider the objective function $\mathcal{L}_n(\theta)$ where the expectations in $\mathcal{L}(\theta)$ are replaced by a sample average over $n$ samples. Let $\hat{\theta}_n = \argmin_\theta \mathcal{L}_n(\theta)$.
\begin{proposition} \label{prop:thetahatn-convergence}
If (i) $I$ is positive definite and (ii) $\sup_\theta |\mathcal{L}_n(\theta) - \mathcal{L}(\theta)| \xrightarrow{p} 0$, then $\hat{\theta}_n \xrightarrow{p} \hat{\theta}$.
\end{proposition}
\begin{proof}
By Corollary \ref{cor:unique-minimizer}, condition (i) ensures that $\hat \theta = \argmin_\theta \mathcal{L}(\theta)$ is a unique minimizer, and hence that changing $\hat \theta$ by a small amount will increase the cost function $\mathcal{L}(\theta)$. Together with the technical condition (ii) on the uniform convergence of $\mathcal{L}_n(\theta)$ to  $\mathcal{L}(\theta)$, this allows one to prove that $\hat{\theta}_n$ converges in probability to $\hat \theta$ as the sample size $n$ increases, following exactly the same reasoning as e.g. in proofs for consistency of maximum likelihood estimation \citep[Section 9.13]{Wasserman2004} or noise-contrastive estimation \citep[Appendix A.3.2]{Gutmann2012a}.
\end{proof}
\begin{corollary}
If (i) $I$ is positive definite, (ii) $\sup_\theta |\mathcal{L}_n(\theta) - \mathcal{L}(\theta)| \xrightarrow{p} 0$, and (iii) there is a parameter value $\theta^\ast$ such that $P^\ast(Y|x)=p(Y|x; \theta^\ast)$, then $\hat{\theta}_n \xrightarrow{p} \theta^\ast$
\end{corollary}
\begin{proof}
With Proposition \ref{prop:thetahatn-convergence}, condition (i) and (ii) ensure that $\hat{\theta}_n$ converges to $\hat{\theta} = \argmin_\theta \mathcal{L}(\theta)$. With Lemma \ref{le:minimiser-characterisation}, $\hat{\theta}$ is also minimizing $\mathbb{E}_{p_x(x)} \text{KL}\left(P^\ast(Y|x) \mid\mid P(Y|x; \theta)\right)$. Hence, if condition (iii) holds, $\hat{\theta}=\theta^\ast$, and the result follows.
\end{proof}
\begin{proposition}[Consistency of the ratio estimator]
If (i) $I$ is positive definite, (ii) $\sup_\theta |\mathcal{L}_n(\theta) - \mathcal{L}(\theta)| \xrightarrow{p} 0$, (iii) $P^\ast(Y|x)=p(Y|x; \theta^\ast)$ for some parameter value $\theta^\ast$, and (iv) the mapping from $\theta$ to $h_\theta^c$ is continuous, then  
\begin{align}
h^i_{\hat \theta_n}(x)- h^j_{\hat \theta_n}(x) \xrightarrow{p} \log\frac{p_i(x)}{p_j(x)}
\end{align}
for all $x$ where $p_x(x)=\sum_c \pi_c p_c(x) >0$.
\end{proposition}
\begin{proof}
By Proposition \ref{prop:thetahatn-convergence}, condition (i) and (ii) ensure that $\hat{\theta}_n$ converges to $\hat{\theta} = \argmin_\theta \mathcal{L}(\theta)$. By Corollary \ref{cor:estimator-parametric}, condition (iii) ensures that $h^i_{\hat \theta}(x)- h^j_{\hat \theta}(x) = \log\frac{p_i(x)}{p_j(x)}$ for all $x$ where $p_x(x)=\sum_c \pi_c p_c(x) >0$. Since continuous functions are closed under addition, the mapping from $\theta$ to $h^i_{\hat \theta}(x)- h^j_{\hat \theta}(x)$ is continuous if condition (iv) holds. We can then apply the continuous mapping theorem to conclude that $h^i_{\hat \theta_n}(x)- h^j_{\hat \theta_n}(x) \xrightarrow{p} h^i_{\hat \theta}(x)- h^j_{\hat \theta}(x)=\log\frac{p_i(x)}{p_j(x)}$, which establishes the result.  
\end{proof}
}

\section{Constructing $M$}
\label{app:constructing_m}

We here elaborate on the three types of auxiliary distributions that we used in this work.

\paragraph{Overlapping Distribution: }
The \cob estimator, $\log \frac{p}{q} = \log \frac{p}{m} - \log \frac{m}{q}$ is defined when $p << m$ and $q << m$. Therefore, $m$ needs to be such that its support contains the supports of $p$ and $q$. Any distribution with full support such as the normal distribution trivially satisfies this requirement. However, satisfying this requirement does not guarantee empirical overlap of the distributions $p$, $q$ with $m$ in finite sample setting. In order to ensure overlap of samples between the two pairs of distributions we recommend the following:
\begin{itemize}
    \item Heavy-tailed Distributions: Distributions such as, Cauchy and Student-t are better choice for $M$ compared to the normal distribution. This is because their heavier tails allow for easily connecting $p$ and $q$ with higher sample overlap when they are far apart (especially in the case of FOD).
    \item Mixtures: Another way to connect $p$ and $q$ using $m$ such that they have their samples overlap, is to use the mixture distribution. Here, we first convolve $p$ and $q$ with a standard normal and then take equal mixtures of the two.
    \item Truncated Normal: If $p$ and $q$ have finite support, one can also use a truncated normal distribution or a uniform distribution that at least spans over the entire support of $q$. This is assuming that $p << q$.
\end{itemize}

\paragraph{Linear Mixing: }
In this construction scheme, distribution $M$ is defined as the empirical distribution of the samples constructed by linearly combining samples $X_p = \{x_p^i\}_{i=1}^N$ and $X_q = \{x_q^i\}_{i=1}^N$ from distributions $p$ and $q$ respectively. That is, $m$ is the empirical distribution over the set $X_m = \{x_m^i | x_m^i = \alpha x_p^i + (1-\alpha)x_q^i, x_p \in X_p, x_q^i \in X_q \}_{i=1}^N$, where $\alpha$ is not constrained to create a convex mixture. This construction is related to the linear combination auxiliary of \cite{tre}. In TRE, the auxiliary distribution is defined as the empirical distribution of the set $X_m = \{x_m^i | x_m^i = \sqrt{1-\alpha^2} x_p^i + \alpha x_q^i,  x_p \in X_p, x_q^i \in X_q \}_{i=1}^N$, where $0 \leq \alpha \leq 1$. This weighting scheme skews the samples from the auxiliary distribution towards $p$. Therefore, care needs to be taken when $p$ and $q$ are finite support distributions so that the samples from the auxiliary distributions do not fall out of the support of $q$.

Using either of the weighting schemes, one can construct $K$ different auxiliary distributions. \cob can either use these $K$ auxiliary distributions separately using a K+2-way classifier or define a single mixture distribution using them as component distributions and train a 3-way classifier. We refer to this construction as Mixture of Linear Mixing.

\paragraph{Dimension-wise Mixing: }In this construction scheme, that is borrowed from TRE as it is, $M$ is defined as the empirical distribution of the samples generated by combining different subsets of dimensions from samples from $p$ and $q$. We describe the exact construction scheme from TRE below for completeness: 

Given a $d$-length vector $x$ and that $d$ is divisible by $l$, we can write down $x = (x[1],...x[l])$, where each $x[i]$ has length $d/l$. Then, a sample from the $k$th auxiliary distribution is given by: $x^i_k = (x^i_q[1], ... x^i_q[j], x^i_p[j+1], ..., x^i_p[l])$, for $j = 1, ..., l)$, where $x^i_p \sim p$ and $x^i_q \sim q$ are randomly paired.

%\section{Appendix}
\section{1D density ratio estimation task}
\label{app:1d_viz}
In Section \ref{sec:1dexp}, we studied three cases in which the two distributions $p$ and $q$ are separated by both FOD and HOD. In all of these 1D experiments, all models were trained with 100,000 samples, and all results are reported across 3 runs with different random seeds. Additionally, we found that \cob worked equally well for 1K and 10K samples. The experimental configurations, including the auxiliary distributions for $\cob$, are detailed in Table \ref{tab:app_1dtask_configs}.

\begin{table*}[ht!]
\centering
\begin{tabular}{@{}llllll@{}}
\toprule
$p$ & $q$ & True KL & \cob @ 1K & \cob @ 10K & \cob @ 100K \\ \midrule
$\mathcal{N}$(-1, 0.08) & $\mathcal{N}$(2, 0.15) & 200.27 & 195.05 & 196.50 & 203.32 \\
$\mathcal{N}$(-2, 0.08) & $\mathcal{N}$(2, 0.15) & 355.82 & 346.92 & 348.97 & 360.35 \\ \bottomrule
\end{tabular}
\caption[]
{\small \cob on 1D density ratio estimation for three settings of sample sizes. \cob estimates the density ratio well for all the three settings.} 
\label{tab:1dtask_app}
\end{table*}

% Please add the following required packages to your document preamble:
% \usepackage{booktabs}
\begin{table*}[]
\centering
\begin{tabular}{@{}llll@{}}
\toprule
$p$            & $q$           & TRE $p_k$    & \cob $m$                                                                                            \\ \midrule
% $\mathcal{N}(0, 1e-6)$     & $\mathcal{N}(0, 1)$    & \begin{tabular}[c]{@{}l@{}}Linear Mixing with \\ $\alpha$ = [6.10e-05, 0.0078, 0.13]\end{tabular}                                                                                                                                                                    & \begin{tabular}[c]{@{}l@{}}Mixture of Linear Mixing with \\ $\alpha$ = [6.10e-05, 0.0078, 0.13]\end{tabular} \\
$\mathcal{N}(-1, 0.08)$ & $\mathcal{N}(2, 0.15)$ & \begin{tabular}[c]{@{}l@{}}Linear Mixing with \\ $\alpha$ = [0.053, 0.11, \\ 0.16, 0.21, 0.26, 0.31, \\ 0.37, 0.42, 0.47, 0.53, \\ 0.58, 0.63, 0.68, 0.74, \\ 0.79, 0.84, 0.89, 0.95]\end{tabular}                                                                   & $\mathcal{C}(0,1)$                                                                                           \\
$\mathcal{N}(-2, 0.08)$ & $\mathcal{N}(2, 0.15)$ & \begin{tabular}[c]{@{}l@{}}Linear Mixing with \\ $\alpha$ = [0.03, 0.07, \\ 0.1, 0.14, 0.17, 0.21, \\ 0.24, 0.28, 0.31, 0.34, \\ 0.38, 0.41, 0.45, 0.48, \\ 0.52, 0.55, 0.59, 0.62, \\ 0.66, 0.69, 0.72, 0.76, \\ 0.79, 0.83, 0.86, 0.9, \\ 0.93, 0.97]\end{tabular} & $\mathcal{C}(0,1)$ \\

$\mathcal{N}(-10, 1)$    & $\mathcal{N}(10, 1)$    & \begin{tabular}[c]{@{}l@{}}Linear Mixing with \\ $\alpha$ = [0.11, 0.22, \\ 0.33, 0.44, 0.55, 0.66, \\ 0.77, 0.88]\end{tabular} & $\mathcal{C}(0,2)$ \\
 \bottomrule
\end{tabular}
\caption[]
{\small Experiment configurations for Table \ref{1dtask} of main text and Table \ref{tab:1dtask_app} of Appendix \ref{app:1d_viz}}
\label{tab:app_1dtask_configs}
\end{table*}
% \begin{table}[]
% \centering
% \begin{tabular}{@{}lllllll@{}}
% \toprule
% \textbf{$p$}            & \textbf{$q$}           & \textbf{GT-KL} & \textbf{BC-DRE} & \textbf{TRE} & \textbf{F-DRE} & \textbf{\cob} \\ \midrule
% $\mathcal{N}(0, 1)$     & $\mathcal{N}(0, 1)$    & 13.32          & 1.82                   & 13.05        & 4.57           & 13.32                        \\
% $\mathcal{N}(-5, 1)$    & $\mathcal{N}(5, 1)$    & 50             & 69.56                 & 34.51        & 15.96          & 53.97                        \\
% $\mathcal{N}(-1, 0.08)$ & $\mathcal{N}(2, 0.15)$ & 200.27         & 19.2                 & 142.64       & 15.84          & 201.32                       \\
% $\mathcal{N}(-2, 0.08)$ & $\mathcal{N}(2, 0.15)$ & 355.82         & 17.95                  & 227.98       & 14.49          & 358.81                       \\ \bottomrule
% \end{tabular}
% \caption[]
% {\small 1D density ratio estimation task.} 
% \label{tab:app_1dtask}
% \end{table}

\begin{figure*}[ht!]
    \centering
    \begin{subfigure}[b]{0.45\textwidth}
        \includegraphics[width=\textwidth]{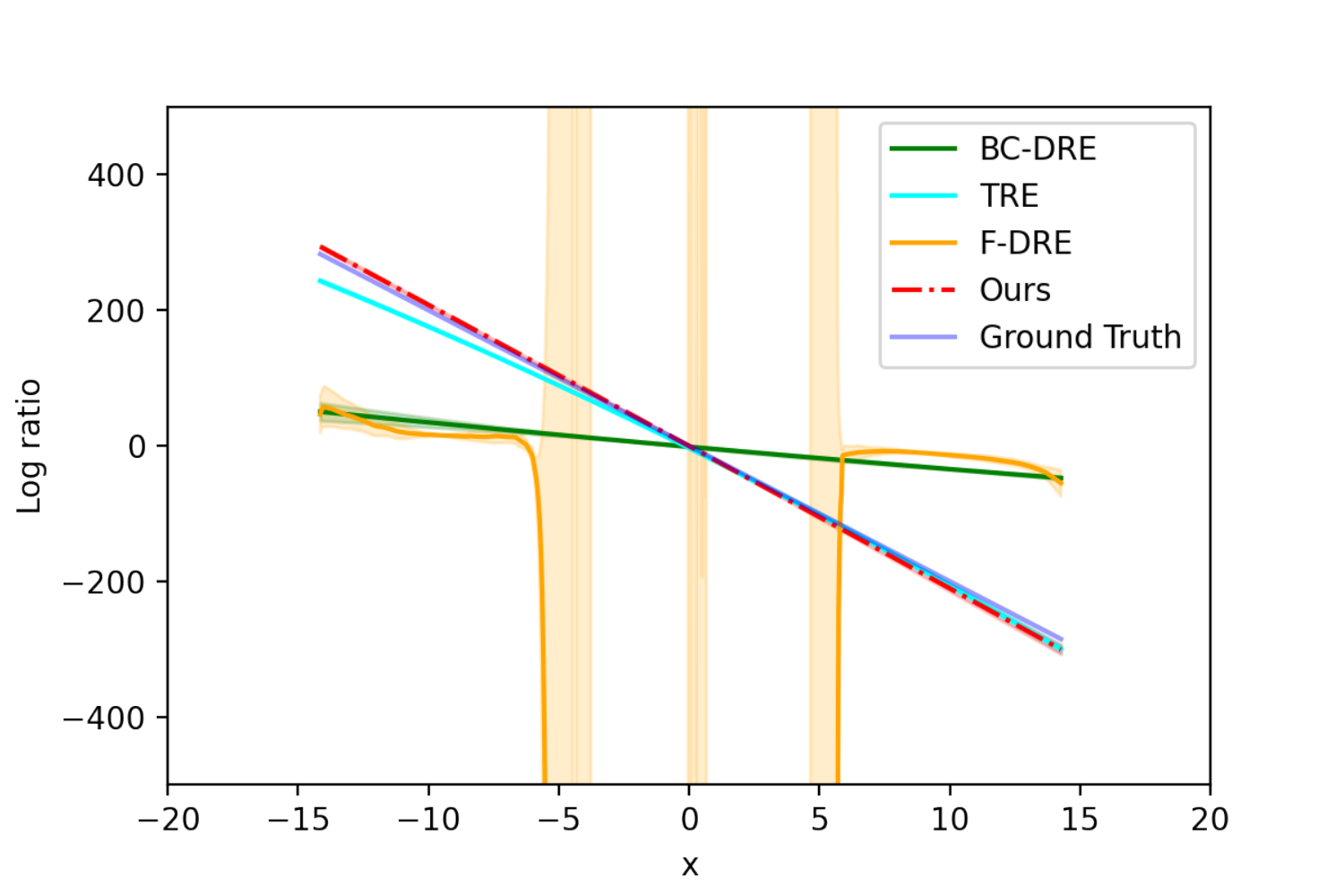}
        \caption{In-domain: $p=\mathcal{N}(-10,1)$ and $q=\mathcal{N}(10,1)$}
        \label{fig:1d_ratios_-10_10_indom}
    \end{subfigure}
    \begin{subfigure}[b]{0.45\textwidth}
        \includegraphics[width=\textwidth]{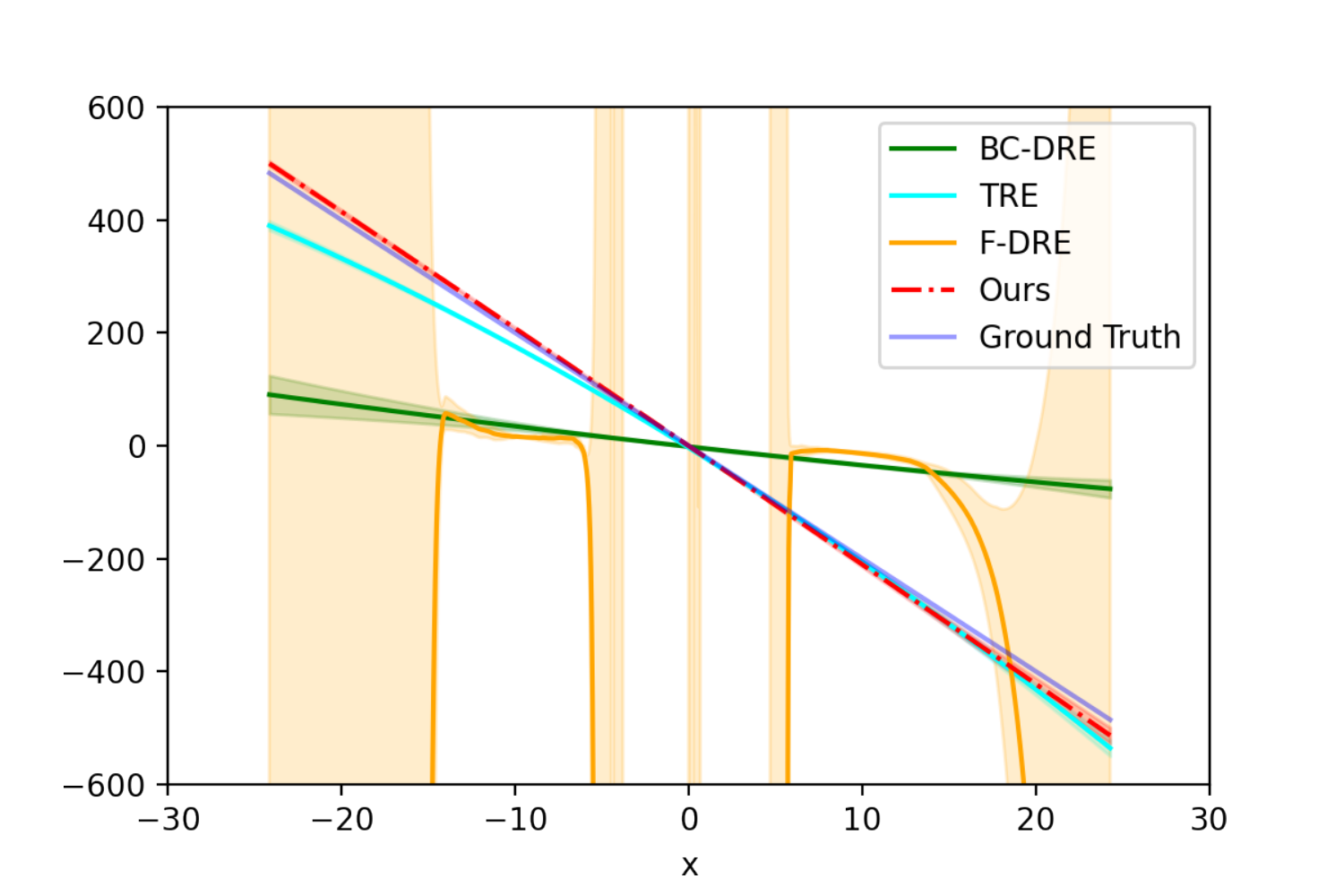}
        \caption{Expanded domain: $p=\mathcal{N}(-10,1)$ and $q=\mathcal{N}(10,1)$}
        \label{fig:1d_ratios_-10_10_outdom}
    \end{subfigure}
    % \vskip\baselineskip
    % \centering
    % \begin{subfigure}[b]{0.45\textwidth}
    %     \includegraphics[width=\textwidth]{fig/1d/mu--1_2_indomain.pdf}
    %     \caption{In-domain: $p=\mathcal{N}(-1,0.08)$ and $q=\mathcal{N}(2,0.15)$}
    %     \label{fig:1d_ratios_-1_2_indom}
    % \end{subfigure}
    % \begin{subfigure}[b]{0.45\textwidth}
    %     \includegraphics[width=\textwidth]{fig/1d/mu--1_2_outdomain.pdf}
    %     \caption{Expanded domain:$p=\mathcal{N}(-1,.08)$ and $q=\mathcal{N}(2,.15)$}
    %     \label{fig:1d_ratios_-1_2_outdom}
    % \end{subfigure}
    % \vskip\baselineskip
    % \centering
    % \begin{subfigure}[b]{0.45\textwidth}
    %     \includegraphics[width=\textwidth]{fig/1d/mu-2_2_indomain.pdf}
    %     \caption{In-domain: $p=\mathcal{N}(-2,0.08)$ and $q=\mathcal{N}(2,0.15)$}
    %     \label{fig:1d_ratios_-2_2_indom_app}
    % \end{subfigure}
    % \begin{subfigure}[b]{0.45\textwidth}
    %     \includegraphics[width=\textwidth]{fig/1d/mu-2_2_outdomain.pdf}
    %     \caption{Expanded domain: $p=\mathcal{N}(-2,0.08)$ and $q=\mathcal{N}(2,0.15)$}
    %     \label{fig:1d_ratios_-2_2_outdom_app}
    % \end{subfigure}
    \caption{1D density ratio estimation analysis. Figure \ref{fig:1d_ratios_-10_10_indom} evaluates the log density ratios on uniform samples inside the domain of the respective training distribution. Figure \ref{fig:1d_ratios_-10_10_outdom} evaluates the log density ratios on uniform samples from an expanded domain of the training distribution. The shading represents 1 standard deviation of the estimates. } \label{fig:1d_ratios_plot_app}
\end{figure*}

\section{Uncertainty Quantification of \cob Log-ratio Estimates with Hamiltonian Monte Carlo}
\label{app:hmc}
\begin{figure*}[]
        \centering
        \begin{subfigure}[b]{0.45\textwidth}
            \centering
            \includegraphics[width=\textwidth]{fig/hmc/uncertainty-ratio-flip=false.pdf}
            \caption[]%
            {\label{fig:ratio_hmc}{\small $\log\frac{p}{q}$ for $p=\mathcal{N}(-1.0,0.1), q=\mathcal{N}(1.0,0.2)$}}
        \end{subfigure}
        \hfill
        \begin{subfigure}[b]{0.45\textwidth}
            \centering
            \includegraphics[width=\textwidth]{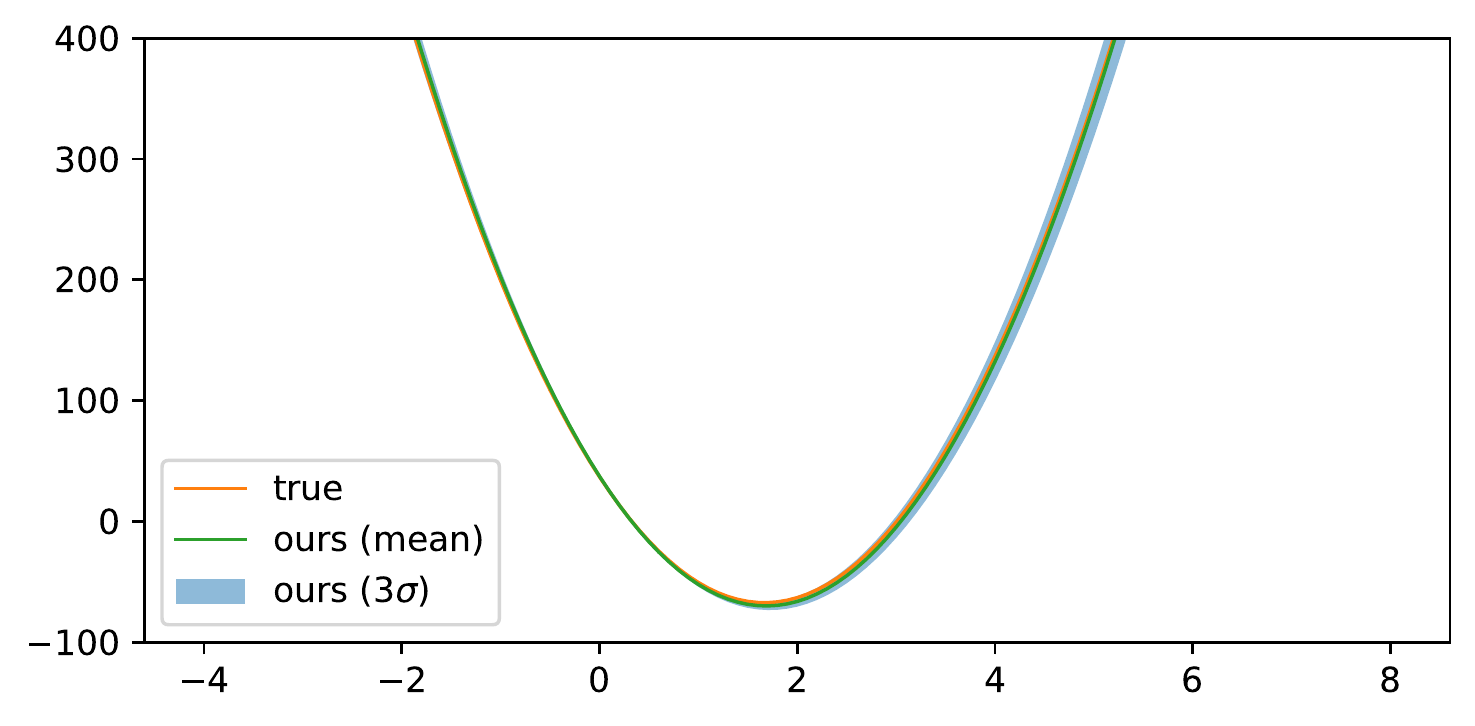}
              \caption[]%
              {\label{fig:ratio_hmc_flip}{\small $\log\frac{p}{q}$ for $p=\mathcal{N}(-1.0,0.2), q=\mathcal{N}(1.0,0.1)$}}
        \end{subfigure}
        \vfill
        \begin{subfigure}[b]{0.45\textwidth}
            \centering
            \includegraphics[width=\textwidth]{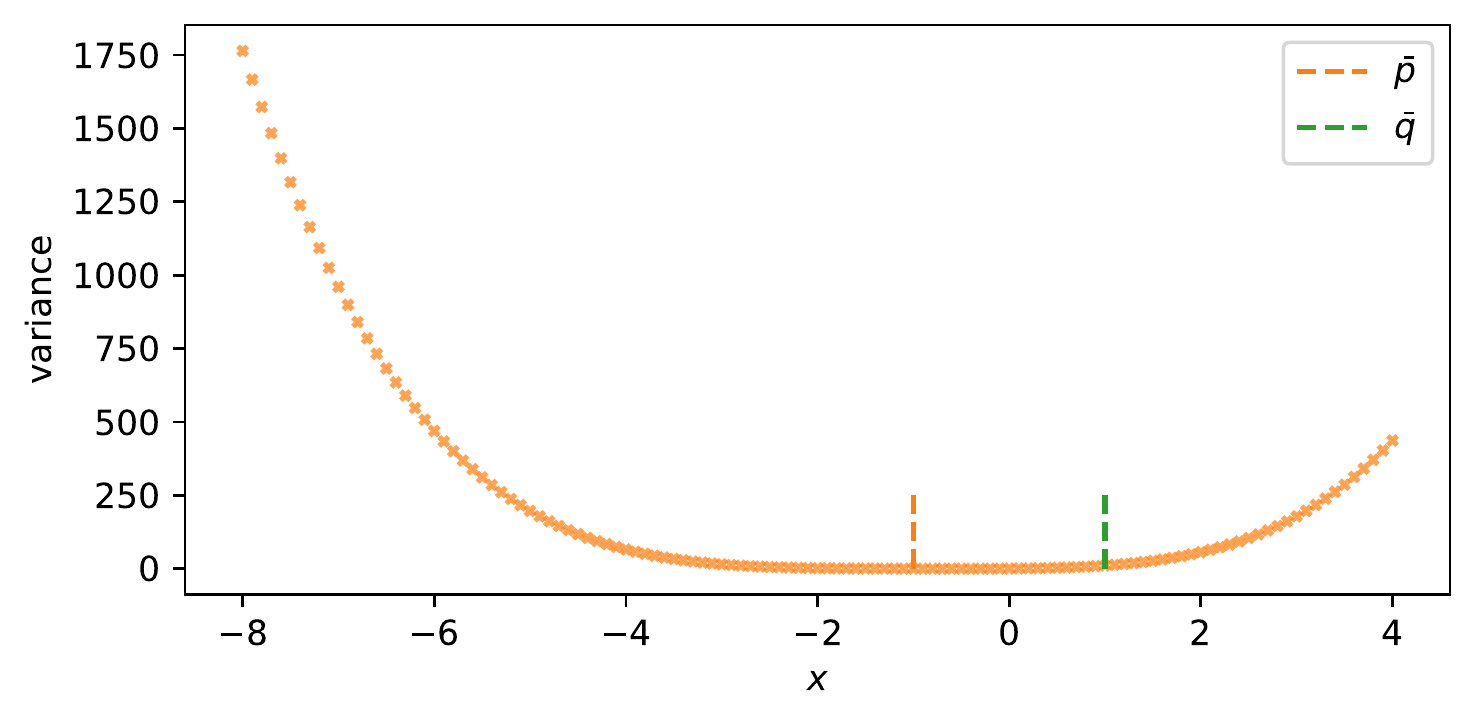}
            \caption[]%
            {\label{fig:var_hmc}{\small Variance of $\log\frac{p}{q}$ as a function of distance from the modes of $p$ and $q$ for $p=\mathcal{N}(-1.0,0.1), q=\mathcal{N}(1.0,0.2)$}}
        \end{subfigure}
        \hfill
        \begin{subfigure}[b]{0.45\textwidth}
            \centering
            \includegraphics[width=\textwidth]{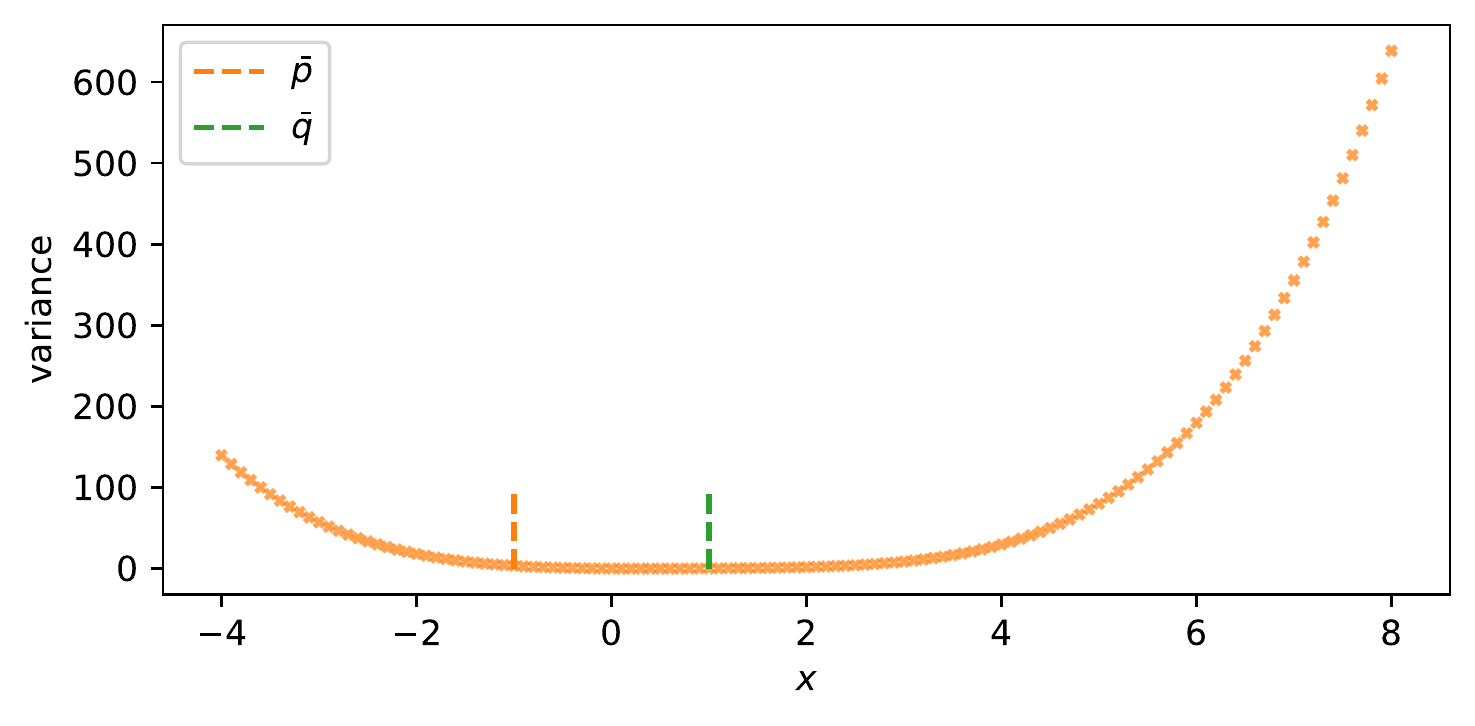}
            \caption[]%
            {\label{fig:var_hmc_flip}{\small Variance of $\log\frac{p}{q}$ as a function of distance from the modes of $p$ and $q$ for $p=\mathcal{N}(-1.0,0.2), q=\mathcal{N}(1.0,0.1)$}}
        \end{subfigure}
        \vfill
        % \vskip\baselineskip
        \caption[]
        {\small Uncertainty quantification for \cob estimator. We plot the 3x standard deviation around the mean in light blue. In plots (c) and (d) the bars show the means of $p$ and $q$.}
        \label{fig:hmc}
    \end{figure*}
    
In the 1D experiments, \cob consistently led to highly accurate KL divergence estimates even in challenging settings where state-of-the-art methods fail. 
To understand why \cob gives such accurate KL estimates, 
we conduct an analysis on the reliability of its log-ratio estimates by analyzing the distribution of the estimates in a Bayesian setup, 
and study how it impacts the KL divergence estimation.
For this analysis, we use a classifier with the standard normal distribution as the prior on its parameters.
% We treat the original objective of \cob as the likelihood term and thus the posterior of the parameters is proportional to ???.
The distribution of the log-ratio estimates is simply the distribution of the estimates from the classifiers with different posterior parameters, which are sampled.
We consider two setups where first we set $p=\mathcal{N}(-1.0,0.1)$ and $q=\mathcal{N}(1.0,0.2)$ and then swap their scales, i.e. $p=\mathcal{N}(-1.0,0.1)$ and $q=\mathcal{N}(1.0,0.2)$. In both the cases, we draw samples from the posterior using an Hamiltonian Monte Carlo (HMC) sampler initialized by the maximum likelihood estimate of the classifier parameter.
We then compute a set of samples of the log-ratio estimates from \cob and estimate the mean and standard deviation using these samples. Figure \ref{fig:hmc} (a) and (b) shows these results. 
We find that \cob is accurate and manifests lowest uncertainty around the region between the means of $p$ ($-1.0$) and $q$ ($+1.0$). The uncertainty increases as we move away from the modes of distributions $p$ and $q$. This is shown in plots (c) and (d), where we plot the variance of the estimates as a function of the location of the sample.

Since KL divergence is the expectation of the log-ratio on samples from $p$ and the high density region of $p$ exactly matches the high confidence region of \cob, it is able to consistently estimate the KL divergence accurately even when $p$ and $q$ are far apart.

%\section{Appendix}
\section{High Dimensional Experiment}
\label{app:high_dim}
\begin{figure*}
    \centering
    \includegraphics[width=\textwidth]{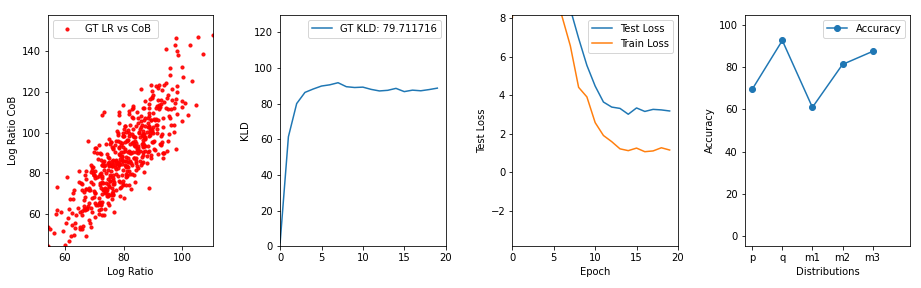}
    \caption{\small{Diagnostic plot for a high dimensional experiment.}}\label{fig:app_highdim_diagnostics}
\end{figure*}
\begin{figure*}
    \centering
    \includegraphics[width=\textwidth]{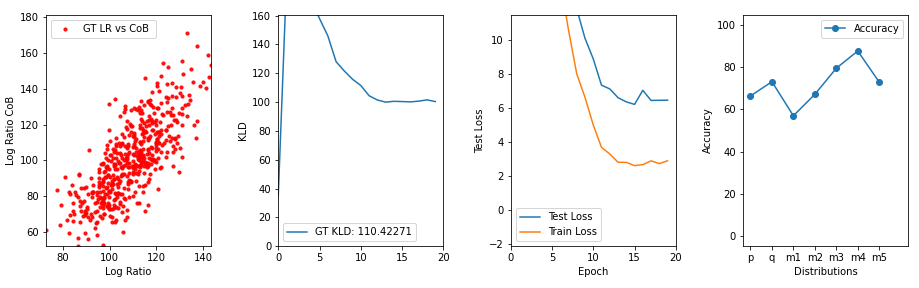}
    \caption{\small{Diagnostic plot for a high dimensional experiment with randomized means.}}\label{fig:app_highdim_diagnostics1}
\end{figure*}
In Section \ref{sec:high_dim}, we showed that \cob performs better than all baseline models when $p$ and $q$ are high dimensional Gaussian distributions. Prior work of \citet{tre} has considered high dimensional cases with HOD only, whereas we additionally consider cases with FOD and HOD to provide a more complete picture. Our results show that \cob outperforms all other methods on the task of MI estimation as the function of the estimated density ratio. It is worth noting that \cob uses only upto 5 auxiliary distributions that are constructed using the linear mixing scheme and beats TRE substantially on cases with both FOD and HOD, although TRE uses upto 15 auxiliary distributions also constructed using linear mixing approach. This demonstrates that our proposal of using the multi-class logistic regression does, in fact, prevent distribution shifts issues of TRE when both FOD and HOD are present and help estimate the density ratio more accurately.

We now describe the \cob configuration and other setup related details. 

\paragraph{Auxiliary Distributions: } For all the high dimensional experiments throughout this work, we construct $m$ using the linear mixing scheme as described in Appendix \ref{app:constructing_m}. Table \ref{tab:apphighdimconfig} provides the number $K$ of auxiliary distributions along with the exact mixing weights for each of the 6 settings. 

\begin{table}[]
\centering
\begin{tabular}{@{}ccc@{}}
\toprule
\textbf{Dim}         & \textbf{MI} & \textbf{$m$ using LM}       \\ \midrule
\multirow{2}{*}{40}  & 20          & {[}0.25,0.5,.75{]}          \\
                     & 100         & {[}0.35,0.5,.85{]}          \\ \midrule
\multirow{2}{*}{160} & 40          & {[}0.25,0.5,.75{]}          \\
                     & 136         & {[}0.15,0.35,0.5,.75,.95{]} \\ \midrule
\multirow{2}{*}{320} & 80          & {[}0.25,0.5,.75{]}          \\
                     & 240         & {[}0.15,0.35,0.5,.75,.95{]} \\ \bottomrule
\end{tabular}
\caption{\small{Configuration of \cob for the high dimensional experiments. LM stands for Linear Mixing} \label{tab:apphighdimconfig}}
\end{table}

As a general principle, we chose these three sets of mixing weights so that their cumulative samples overlap with the samples of $p$ and $q$ similar to how the heavy tailed distribution worked in the 1D case. 
Please note that while heavy tailed distributions can effectively bridge $p$ and $q$ when they have high FOD. However, they do not work as well if the discrepancy is primarily HOD. For example consider $p=\mathcal{N}(0, 1e-6)$ and $q=\mathcal{N}(0,1)$. In this case, setting $m$ to a heavier tailed distribution centered as zero will not be of help. We need $m$ that is concentrated at zero but also maintains a decent overlap with $q$. Linear mixing $p$ and $q$ on the other hand, mixes first and higher order statistics (second or higher) and therefore, populates samples that overlap with both $p$ and $q$. In some cases, we found that a mixture of linear mixing with $K=1$ can also be used to estimate the density ratio. However, this requires using a neural network-based classifier and requires much more tuning of the hyperparameters.

For choosing $K$, we use a grid search based approach. We monitor the classification accuracy across all the $K+2$ distribution. If this accuracy is very high ($>95$\% for all classes), this implies that the classification task is easy and therefore the DRE may suffer from the density chasm issue. On the other hand, if the classification accuracy is too low ($<50$\% for all classes), then again, the DRE does not estimate well. We found that targeting an accuracy curve as shown in Figure \ref{fig:app_highdim_diagnostics} (last panel) empirically leads to accurate density ratio estimation. This curve plots the test accuracy across all the classes and, empirically when it stays between the low and the high bounds of (50\%,95\%), the DRE estimates the ratios fairly well. The first panel shows that \cob estimates the ground truth ratio accurately across samples from all the $K+2$ distributions, the second panel shows that KL estimates of \cob is close to the ground truth KL and the third panel shows that both test and training losses have converged. Figure \ref{fig:app_highdim_diagnostics1} shows another example for the case of randomized means. While \cob also manages to get the ground truth KL correctly and most of the ratio estimates are also accurate, it does, however, slightly overestimate the log ratio for some of the samples from $p$.

% \paragraph{Additional Baseline Comparisons}
% We also compare to \cite{kato2021non} in a high-dimensional setting (row 2 of Table \ref{tab:highdim} with dimensionality of 40 and $\mu_1=-1, \mu_2=1$. In this setting in which the ground truth MI is 100, while sDRE meaningfully estimates the MI as 119.96, the best model from \cite{kato2021non} only estimates it to be $1.60$.

%\section{Appendix}
\section{SpatialMultiOmniglot Experiment}
\label{app:omni}
% \begin{figure*}
% \centering
% \includegraphics[width=0.5\textwidth]{fig/omniglot/omniglot_k.jpeg}
% \caption{Mutual information estimation and representation learning with varying K.\label{fig:omni_k}}
% \end{figure*}

\begin{figure*}
    \centering
    \begin{subfigure}[b]{0.4\textwidth}
        \includegraphics[width=\textwidth]{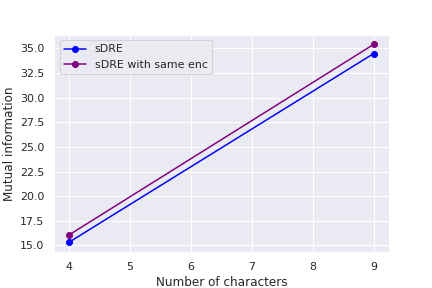}
        \caption{Mutual information estimation}
        \label{fig:omni-mi-sameenc}
    \end{subfigure}
    \begin{subfigure}[b]{0.4\textwidth}
        \includegraphics[width=\textwidth]{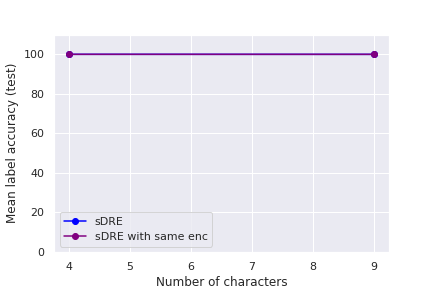}
        \caption{Representation learning accuracy}
        \label{fig:omni-acc-sameenc}
    \end{subfigure}
    \caption{SpatialMultiOmniglot representation learning results with same encoder for $f$ and $g$.}\label{fig:omni_sameenc}
\end{figure*}

SpatialMultiOmniglot is a dataset of paired images $u$ and $v$, where $u$ is a $n \times n$ grid of Omniglot characters from different Omniglot alphabets and $v$ is a $n \times n$ grid containing the next characters of the corresponding characters in $u$. In this setup, we treat each grid of $n \times n$ as a collection of $n^2$ categorical random variables, not the individual pixels. The mutual information $I(u,v)$ can be computed as: $I(u,v)=\sum_{i=1}^{n^2} \log l_i$, where $l_i$ is the alphabet size for the $i^{th}$ character in $u$. This problem allows us to easily control the complexity of the task since increasing $n$ increases the mutual information. 

For the model, as in TRE, we use a separable architecture commonly used in MI-based representation learning literature and model the unnormalized log-scores with functions of the form $g(u)^T Wf(u)$, where $g$ and $f$ are 14-layer convolutional ResNets \citep{he2015resnet}. We construct the auxiliary distributions via dimension-wise mixing---exactly the way that TRE does.

To evaluate the representations after learning, we adopt a standard linear evaluation protocol to train a linear classifier on the output of the frozen encoder $g(u)$ to predict the alphabetic index of each character in the grid $u$.

\subsection*{Additional Experiments}

In addition to the experiments in the main text, we run an additional experiment with SpatialMultiOmniglot to test the effect of using the same encoder for $g$ and $f$ (i.e, modeling the unnormalized log-scores with the form $g(u)^TWg(v))$ instead of $\log p(u,v)=g(u)^TWf(v))$.

\paragraph{Single Encoder Design:}We test the contribution of using two different encoders $f$ and $g$ instead of one. As seen in Figure \ref{fig:omni_sameenc}, in both cases of $d=4, 9$, the two models reach slightly different but similar MI estimates, but, interestingly, do not differ at all in the test classification accuracy. Empirically, we also found that using one encoder helps the model converge to much faster. Overall, this experiment demonstrates that using two different encoders does not necessarily work to our advantage.

\section{TRE on Finite Support Distributions for $K=1$}
\label{app:trevscob}

% \begin{figure*}[]
%         \centering
%         \begin{subfigure}[b]{\textwidth}
%             \centering
%             \caption[]%
%             {{\small TRE on $p=\mathcal{N}(-1,0.1)$ and $q=\mathcal{N}(1,0.2)$ with TRE's skewed linear auxiliary distributions}}
%             \includegraphics[width=\textwidth]{fig/1d/1D_demo_tre_K=3_from_logpq.png}
%             \label{fig:app_tre_insight}
%         \end{subfigure}
%         % \hfill
%         \begin{subfigure}[b]{\textwidth}
%             \centering
%             \caption[]
%             {\small \cob on $p=\mathcal{N}(-1,0.1)$ and $q=\mathcal{N}(1,0.2)$ using $m=\cauchydist(0,1)$ as the auxiliary distribution.}
%             \includegraphics[width=\textwidth]{fig/1d/1D_demo_cob_K=1_cauchy_logpq.png}
%             \label{fig:app_1dcobdemo}
%         \end{subfigure}
        
%         \begin{subfigure}[b]{\textwidth}
%             \centering
%             \caption[]%
%             {{\small \cob using TRE auxiliary distributions.}}
%             \includegraphics[width=\textwidth]{fig/1d/1D_demo_cob_K=3_logpq.png}
%             \includegraphics[width=\textwidth]{fig/1d/1D_cob_trewm_pm_qm.png}
%             \label{fig:app_cob_insight_2}
%         \end{subfigure}
%         % \vskip\baselineskip
%         \caption[]
%         {\small TRE vs \cob on $p=\mathcal{N}(-1,0.1)$ and $q=\mathcal{N}(1,0.2)$ using 3 intermediate distributions $p_1, p_2, p_3$ constructed using the \textit{linear-combination} construction.}
%         \label{fig:app_trevscob}
%     \end{figure*}

For $K=1$, TRE proposes the following telescoping: $\log p/q = \log p/m + \log m/q$. As such, for TRE to be well defined, $\frac{dM}{dQ}$ i.e. the Radon-Nikodym Derivatives (RND) needs to exist. The consequence of this is that TRE is only defined when $p << m << q$. However this condition easily breaks if, for example, $p$ and $q$ are mixtures of finite support distributions except for the trivial case when support of $m$ is exactly equal to the support of $q$. 

\begin{figure*}[ht!]
        \centering
        \begin{subfigure}[b]{.45\textwidth}
            \centering
            \caption[]%
            {{\small Setup}}
            \includegraphics[width=0.9\textwidth]{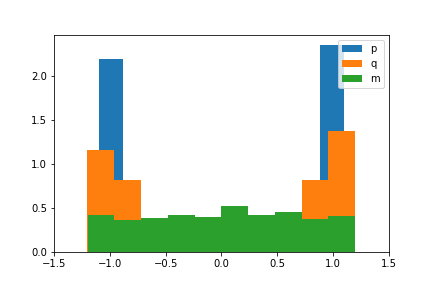}
            \label{fig:hist_mix}
        \end{subfigure}
        \begin{subfigure}[b]{.45\textwidth}
            \centering
            \caption[]
            {\small $dM/dQ$}
            \includegraphics[width=0.9\textwidth]{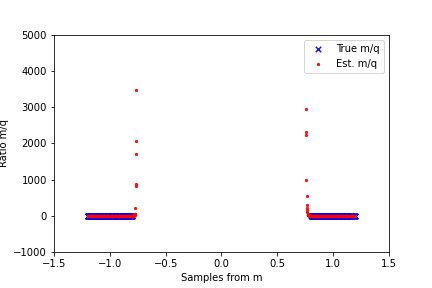}
            \label{fig:1dtre_mix}
        \end{subfigure}
        % \begin{subfigure}[b]{.3\textwidth}
        %     \centering
        %     \caption[]
        %     {\small $dQ/dM$}
        %     \includegraphics[width=0.9\textwidth]{fig/1d/mixture_truncated_q_m.png}
        %     \label{fig:1dcob_mix}
        % \end{subfigure}
        \caption{$dM/dQ$ on mixtures of distributions with finite support\label{fig:mix_tre_cob}}
    \end{figure*}

We now demonstrate this with a specific example in Figure \ref{fig:mix_tre_cob}. Here we set $p = 0.5 \times \mathcal{TN}(-1, 0.1, low=-1.1, high=-0.9) + 0.5 \times \mathcal{TN}(1, 0.1, low=0.9, high=1.1)$ and 
$q = 0.5 \times \mathcal{TN}(-1, 0.2, low=-1.2, high=0.8) + 0.5 \times \mathcal{TN}(1, 0.2, low=0.8, high=1)$, as shown in Figure \ref{fig:hist_mix} where $\mathcal{TN}$ stands for Truncated Normal distribution. 
We set the auxiliary distribution $m$ in TRE to $m = \mathcal{TN}(0, 1, low=-1.2, high=1.2)$ using the proposed \emph{overlapping distribution} construction. As such, $p << q << m$ and therefore, TRE is undefined for the second term as $\frac{m}{q}$ is not defined for samples from $m$ that are outside the support of $q$. It can be clearly seen in Figure \ref{fig:1dtre_mix} that $\frac{m}{q}$ blows up to very high values on samples from $m$ where $q$ does not have any support. Similar examples can be constructed for all the auxiliary distribution construction schemes proposed in \citet{tre}.

Please note, despite $dM/dQ$ being undefined, when used with the proposed $m$, TRE estimates $\log r_{p/q}$ accurately on samples from $p$. We conjecture that this is because, the numerical estimation of the $dM/dQ$ is finite over the support of $p$.

\end{document}